\documentclass[twocolumn]{fairmeta} % option "twocolumn" available
\usepackage{lipsum}

\usepackage{graphicx}
\usepackage{colortbl}
\usepackage{multirow}
\usepackage{collcell}
\usepackage{xifthen}
\usepackage{tikz}
\usepackage{pgf} 
\usepackage{etoolbox}

\AtBeginEnvironment{tabular}{\footnotesize}
\AtBeginEnvironment{texttt}{\small}

\title{A Picture is Worth More Than 77 Text Tokens: Evaluating CLIP-Style Models on Dense Captions}

\author[*,\dagger]{Jack Urbanek}
\author[*,1,2,3]{Florian Bordes}
\author[1]{Pietro Astolfi}
\author[1]{Mary Williamson}
\author[1]{\\Vasu Sharma}
\author[1,2,4,5]{Adriana Romero-Soriano}

\affiliation[1]{FAIR at Meta}
\affiliation[2]{Mila}
\affiliation[3]{Universite de Montreal}
\affiliation[4]{McGill University}
\affiliation[5]{Canada CIFAR AI chair}

\contribution[*]{Joint first author}
\contribution[\dagger]{Work done at Meta}

\abstract{\lipsum[1]}

\date{\today}
\correspondence{\email{fbordes@meta.com}}

\metadata[Code]{\url{https://github.com/facebookresearch/DCI}}
\abstract{\begin{abstract}

Curation methods for massive vision-language datasets trade off between dataset size and quality.
However, even the highest quality of available curated captions are far too short to capture the rich visual detail in an image. 
To show the value of dense and highly-aligned image-text pairs, we collect the Densely Captioned Images (DCI) dataset, containing 7805 natural images human-annotated with mask-aligned descriptions averaging above 1000 words each. 
With precise and reliable captions associated with specific parts of an image, we can evaluate vision-language models' (VLMs) understanding of image content with a novel task that matches each caption with its corresponding subcrop. 
As current models are often limited to 77 text tokens, we also introduce a summarized version (sDCI) in which each caption length is limited.
We show that modern techniques that make progress on standard benchmarks do not correspond with significant improvement on our sDCI based benchmark. 
Lastly, we finetune CLIP using sDCI and show significant improvements over the baseline despite a small training set.
By releasing the first human annotated dense image captioning dataset, we hope to enable the development of new benchmarks or fine-tuning recipes for the next generation of VLMs to come.\looseness-1

\end{abstract} }

\begin{document}

\maketitle

\section{Introduction}
\label{sec:intro}

State-of-the-art vision-language models (VLMs) are often trained on large scale datasets such as LAION-400M~\citep{schuhmann2021laion400m}, YFCC100M~\citep{Thomee_2016}, or other undisclosed datasets crawled from the web. These datasets are formed by collecting images from the web and using alt-text (or other local text on the webpage) to create loose image-text pairs. These can then be filtered down trading off on quantity for quality~\citep{sharma-etal-2018-conceptual,radenovic2023filtering}. Still, recent work has demonstrated that throwing these loose captions out entirely in favor of generated captions, with enhanced quality and density, can produce improved results~\citep{doveh2023dense}. Other works~\citep{li2023inverse, li2023scaling, xu2023cit, abbas2023semdedup} have demonstrated that it is possible to get CLIP-level performance using a vastly reduced compute, often by throwing away portions of the data resulting in more balance between image and text modalities. However, those approaches rely on automatic pipelines which do not generate reliable and long captions that can capture rich visual details in an image. From this it appears no existing dataset has high-quality image descriptions that are tightly-coupled enough with the image to train for or evaluate a deep alignment between the two domains.\looseness-1

% Recent VL benchmarks and their limitations
In the absence of high quality captions to evaluate VLMs, benchmarks such as ARO~\citep{aroNegClip} and VL-Checklist~\citep{zhao2023vlchecklist} often complement image-caption pairs with hard negatives that are generated by slightly altering the initial (positive) description.
Progress on these benchmarks has been rooted in training VLMs with negatives of similar construction to the tests \citep{aroNegClip} rendering the methodologies ineffective on datasets such as Winoground~\citep{winoground}. Recent works~\citep{lin2023visualgptscore} have called the evaluation capacity of many of these benchmarks into question, given how effective language-prior-based methods perform. More specifically, given the unlikeliness of the hard negative captions in these benchmarks, a good text encoder can achieve close to 100\% accuracy without looking at the images. Moreover,~\citet{bordes2023pug} have shown that most improvements observed on ARO or VL-Checklist do not translate on simple synthetic benchmarks for which the negative caption is as likely as the positive one. Since the use of VLMs is significantly increasing, it is crucial to make sure that we have a diverse suite of reliable benchmarks to asses their abilities.\looseness-1 %, including over what would be considered standard images

\begin{figure*}
  \centering
  \includegraphics[width=0.9\textwidth]{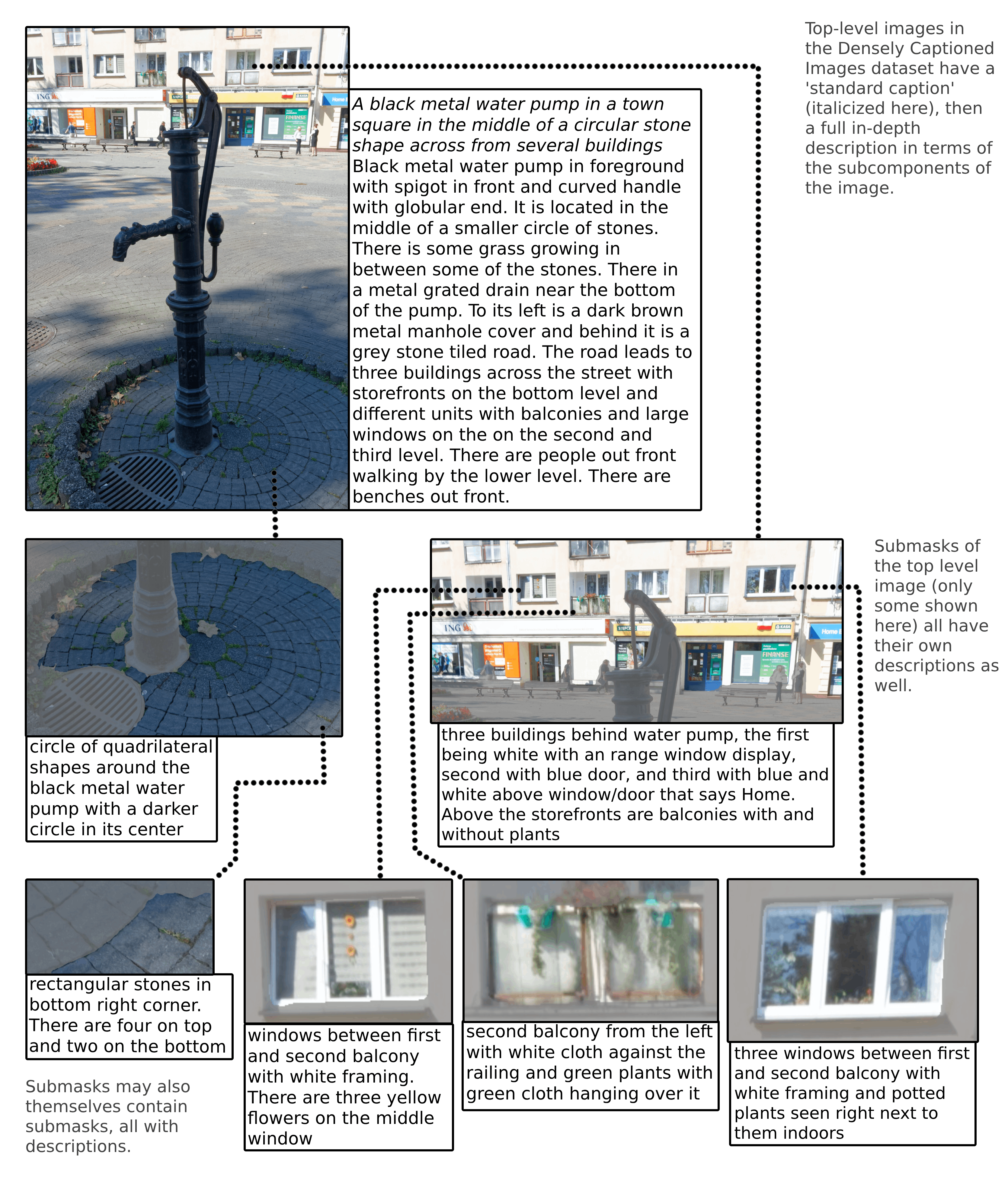}
  \caption{One example from the Densely Captioned Images dataset. Only part of the submask hierarchy is shown.}
  \label{fig:dc_example}
\end{figure*}

In this paper, we introduce the Densely Captioned Images dataset, a collection of 7805 images with dense and mask-aligned descriptions averaging above 1000 words each. One such example is provided in Figure~\ref{fig:dc_example}, displaying just a subset of the collected text paired with their aligned masks. We demonstrate how to leverage this dataset to evaluate VLMs in two ways after summarizing captions to fit into CLIP's 77 token limit, both with a negatives-based test as well as a novel matching task, referred to as \emph{subcrop-caption matching}, that requires selecting appropriate captions for different regions of the same image. We evaluate existing baselines, and observe that no models perform well at both concurrently, and improved performance via negatives-based training comes at the cost of decreased performance on subcrop-caption matching. We also run some experiments using the summarized DCI as a fine-tuning dataset to evaluate the effectiveness of these captions for improving a model's performance on other benchmarks, and compare the efficiency per-example to that from the automated annotation setup in DAC~\citep{doveh2023dense}.\looseness-1

To summarize, our contributions are:
\begin{itemize}
    \item We release the Densely Captioned Images (DCI) dataset, which contains dense and mask-aligned captions, alongside an LLM-summarized version (sDCI) containing captions under 77 tokens for use with current VLMs.
    \item We provide a new benchmark for VLMs based on sDCI to evaluate fine-grained vision-language understanding, and show that no existing model can perform well at matching captions from within one image to corresponding sub-sections of that image.
    \item We show that fine-tuning with high quality image-caption pairs is as good on ARO and VL-Checklist as fine-tuning on at least $10\times$ the automatically annotated data, and that even without utilizing explicit negatives these pairs can improve performance on VL-C-Object from 81.17\% to 88.37\% .\looseness-1
\end{itemize}

\section{Related Works}
\label{sec:related-works}

The massive, loosely-labeled dataset approach that has enabled VLMs like CLIP~\citep{CLIP} and powerful successors like BLIP2~\citep{BLIP2}, Flamingo~\citep{alayrac2022flamingo}, CM3leon~\citep{yu2023scaling}, and many others, has been a clear forward step in vision-language modeling. Still recent benchmarks show that models trained in this manner display clear drawbacks in reasoning skills. Additional techniques have been proposed and adopted recently to close this gap, discussed below.

%-------------------------------------------------------------------------
\paragraph{Vision-Language Datasets.}
Over the last decade, there have been significant dataset collection efforts connecting images and text. Earlier works focused on curating datasets by leveraging human annotations, see \textit{e.g.}, \textbf{COCO}~\citep{chen2015microsoft}, \textbf{Visual Genome}~\citep{krishna2016visual}, and \textbf{Flickr30k}~\citep{young-etal-2014-image}. The process resulted in high quality annotations, which were however oftentimes limited by the caption content -- \textit{i.e.}, relatively short phrases (5.1 to 10.3 words on average) grounded at image level or region level -- and the data annotation scale (30k to 130k images). To increase scale, researchers gathered web-crawled data and introduced large scale datasets such as \textbf{YFCC100M}~\citep{Thomee_2016}, which contains 100M media objects. Yet, crawling the web oftentimes results in little correspondence between image and text pairs. To reduce noise between image and text pairs, efforts such as \textbf{SBU}~\citep{NIPS2011_5dd9db5e} queried Flickr and filtered the noisy results, obtaining a $\sim$1M images. Moreover, \textbf{Conceptual Captions} (CC)~\citep{sharma-etal-2018-conceptual} crawled a dataset of $\sim$12M images and alt-text pairs, and included a protocol to filter noisy text-image pairs, resulting in 3M data points. Relaxing the filtering protocol allows to trade data quality for scale. Crawling alt-text also resulted in relatively short text descriptions with 10.3 words on average, which are most often grounded at image level. \textbf{Localized Narratives}~\citep{ponttuset2020connecting} was introduced as a dense visual grounding dataset leveraging a multi-modal annotation procedure, collecting $\sim$850k text-image pairs with 36.5 words/caption on average. \textbf{RedCaps}~\citep{desai2021redcaps} constituted another effort yielding large scale ($\sim$12M) web-curated data by exploring alternate data sources of high quality data instead of devising complex filtering strategies. \textbf{Wikipedia-based image-text dataset} (WIT)~\citep{wit} extended dataset creation efforts by gathering a multilingual dataset of text-image-pairs consisting of $\sim$11.5M images. \textbf{LAION-5B}~\citep{schuhmann2022laion5b} further increased the web-crawling efforts by gathering a multilingual dataset of text-image pairs, and filtered the collected data with a pre-trained CLIP~\citep{CLIP} model. Following, \textbf{LAION-CAT}~\citep{radenovic2023filtering} reduced noisy examples from LAION-5B by filtering for caption complexity, \textit{i.e.}, captions that do not contain any action, and for text spotting, \textit{i.e.}, images that contain rendered text. \textbf{MetaCLIP}~\citep{xu2023demystifying} has also been released as an open dataset for reproducing CLIP. These very large scale datasets have been successfully used to advance the state-of-the-art of VLMs.\looseness-1
%-------------------------------------------------------------------------
\paragraph{Vision-Language Evaluation Benchmarks.}
Several recent advances in visual-language learning have focused on creating comprehensive benchmarks to evaluate model performance in more holistic ways. These benchmarks are instrumental in pushing the envelope of what VLM can understand and process, ensuring they move beyond superficial image-text matching towards genuine understanding of intricate relationships between visual and linguistic elements. 
In particular, \textbf{VL-CheckList}~\citep{zhao2023vlchecklist} and \textbf{ARO}~\citep{aroNegClip} assess the VLM capabilities beyond average downstream task accuracy, by focusing on a model's ability to understand objects, attributes, order or relations.
ARO's extensive scope, uncovers limitations in VLMs such as poor relational understanding and lack of order sensitivity. %, challenging them to go beyond a 'bags-of-words' approach. 
\textbf{Winoground}~\citep{winoground} tests models for visio-linguistic compositional reasoning by asking VLM to match two images with two captions containing the same set of words but in different orders. This task requires models to discern the meaning conveyed by the order of words, reflecting different visual scenes. Current VLMs perform only marginally better than chance, highlighting a significant gap in compositional reasoning. \textbf{CREPE} (Compositional REPresentation Evaluation)~\citep{crepe} evaluates two aspects of compositionality: systematicity and productivity. Systematicity is measured by the model's ability to represent seen versus unseen atoms and their compositions, while productivity gauges the model's capacity to understand an unbounded set of increasingly complex expressions. Finally, \textbf{PUG} (Photorealistic Unreal Graphics)~\citep{bordes2023pug} uses synthetic data to asses the compositional reasoning abilities of VLMs by progressively increasing the complexity of a given generated scene.
One issue with these evaluation datasets is their frequent reliance on COCO, either directly as in ARO, or through Visual Genome as in VL-Checklist or CREPE. It is difficult to find an evaluation dataset of sufficient scale without COCO. 

\begin{figure*}[ht!]
  \centering
  \includegraphics[width=0.9\textwidth]{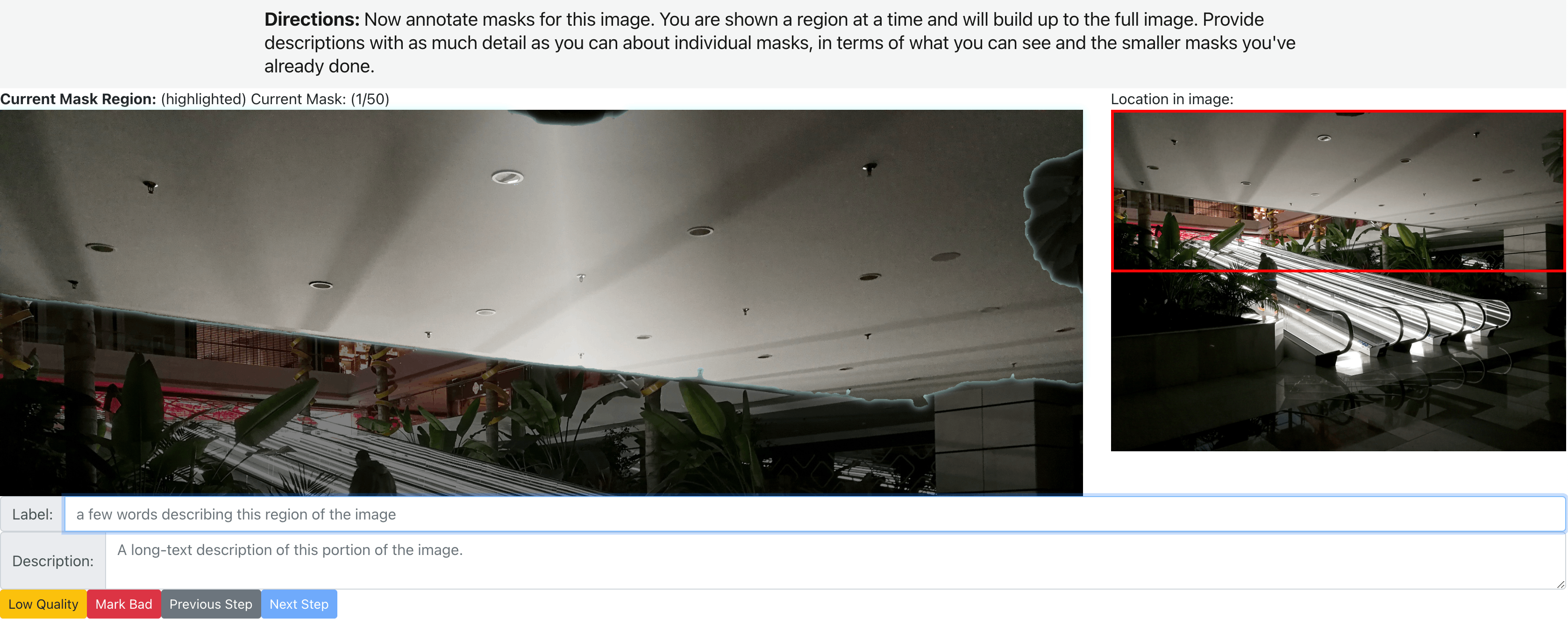}
  \caption{Annotation view for writing description for masks of the image. The masked region appears highlighted for clarity.}
  \label{fig:mask-procedure}
\end{figure*}

%-------------------------------------------------------------------------
\paragraph{Vision-Language models.}
Recent VLM advancements have built upon the foundational work of \textbf{CLIP}~\citep{CLIP}, which leveraged large-scale image-text pairs to jointly pre-train an image encoder and a text encoder to predict which images are paired with which texts in a contrastive learning paradigm. \textbf{NegCLIP} build upon CLIP by leveraging negative captions when training. \textbf{BLIP} (Bootstrapping Language-Image Pre-training)~\citep{BLIP} uses a new framework that bootstraps the captions from noisy web data for both understanding and generation tasks. Its successor \textbf{BLIP-2}~\citep{BLIP2} further streamlines the process by utilizing off-the-shelf frozen pre-trained image encoders and language models, bridging the modality gap with a lightweight querying mechanism. \textbf{Clip-rocket}~\citep{fini2023improved} improves VLM baselines by showing that applying image and text augmentations makes up for most of the improvement attained by prior VLMs. \textbf{Flava}~\citep{singh2022flava} proposes a foundation VLM model by combining existing VLMs objectives together with auxiliary in-modality losses for the text and vision encoders. \textbf{X-VLM}~\citep{xvlm} achieves success with a pretraining method matching sub-portions of the text to regions of the image at multiple granularities. % \textbf{Flamingo}~\cite{alayrac2022flamingo} combines pre-trained vision and language models to converse with users and respond to visual inquiries, setting a new standard for few-shot VLMs. The 80 billion parameter model  excels in few-shot learning, outperforming previous models on 16 vision-language benchmarks.\looseness-1
These models introduces improvements over CLIP, focusing on efficiency, adaptability, and reducing the need for extensive labeled datasets, thereby pushing the boundaries of vision-language pre-training. The closest work to our approach is \textbf{DAC} (Densely Aligned Captions)~\citep{doveh2023dense}, which %enhance compositional reasoning in VLM by improving 
improves with an automated LLM based pipeline the \emph{caption quality} and \emph{density}. % of paired vision-language datasets. %, and with techniques like Multiple Instance Learning and negative text augmentation. 
%By demonstrating that the DAC-enhanced CLIP model exhibits substantial gains on VL-checklist and ARO benchmarks, 
By showing that DAC-enhanced CLIP models exhibit substantial gains on some benchmarks, this work underscores the critical role that caption quality and density play in the efficacy of VLMs. We build on this insight and explore how to further increase the caption quality and density by relying on human annotators, and analyze how that impacts downstream model performance.\looseness-1

\section{Dataset Construction}
\label{sec:dataset-construction}

The Densely Captioned Images dataset, or \textbf{DCI}, consists of 7805 images from SA-1B~\citep{kirillov2023segment}, each with a complete description aiming to capture the full visual detail of what is present in the image. Much of the description is directly aligned to submasks of the image.
An example is shown in Figure~\ref{fig:dc_example}. In the top left we see the full image of a water pump, with an associated description.
The italicized section is collected as a \emph{standard caption}, aiming to summarize the full image in about a sentence, similar to existing caption datasets. The remainder of that first description contains details about the relationship between visible entities in the image, as well as in-depth descriptions of regions that are not described as part of the submasks. All other text describing the image is associated with submasks of the image. Each submask has its own free-text label (not pictured) and description, and may also contain further submasks. Here for instance we see submasks for windows and balconies as being contained in the submask capturing three buildings in the background.

%-------------------------------------------------------------------------
\subsection{Preparation}

\begin{figure*}[ht]
  \centering
  \includegraphics[width=1.0\textwidth]{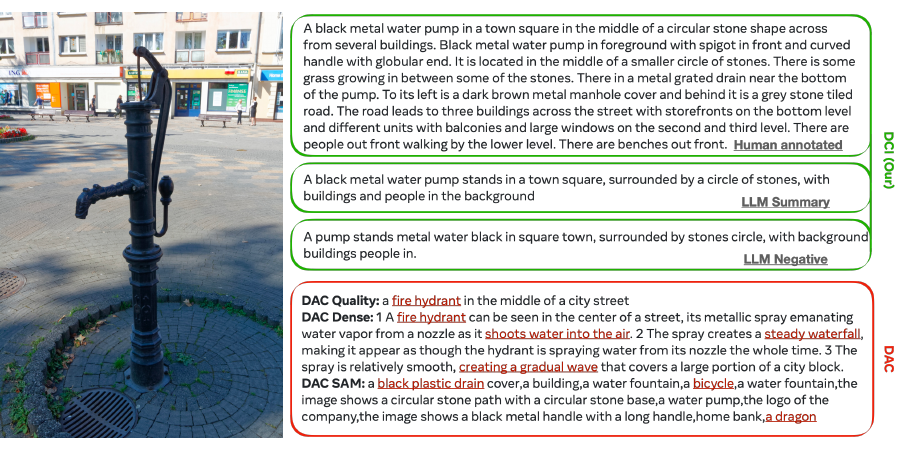}
  \caption{Example of a Llama2-generated summary and negative that comprise sDCI. Each image and submask have multiple summarizations and negatives. We also compare the caption quality between DAC~\citep{doveh2023dense} and DCI. In contrast to DCI that relies on human annotations, DAC used an automatic pipeline based on LLM for captioning. As we observe in this example, the DAC captions can suffer from hallucinations and miss important elements of the photo. In this work we argue that while improving automatic pipeline is an important research direction, for now the captions proposed are not reliable enough to be used to evaluate models and assess their abilities.}
  \label{fig:dac_vs_dci}
\end{figure*}

In order to collect the data, we first select images from a random privacy-mitigated subset of SA-1B. We then procedurally extract subregions of each image to annotate, as we found in initial trials that crowdsourcing both regions and descriptions concurrently overcomplicated the task and successful annotation rate. For this process, we turn to the Segment Anything Model (SAM)~\citep{kirillov2023segment} and adapt their standard method to extract all masks from an image.\looseness-1

For the extraction process, SAM usually relies on a grid of points across the entire image. In order to increase the possibility of selecting interesting regions worth annotating, we additionally apply a canny filter~\citep{cannyfilter} and select random points within a radius from discovered edges. We then run SAM to detect all masks using both the grid and the near-edge points. Once the masks are returned, we establish a hierarchy of submasks by thresholding the number of overlapping pixels between two masks to determine if one is a submask of the other, or if the two masks should be joined. This helps reduce some of the noise introduced by the automatic masking process, and leaves us with a tree-like structure for the masks. Lastly, we remove any masks that are too small. \looseness-1
We note that undergoing this process does not result in every detail of each image being selected as a candidate for annotation, and as such instances in the DCI dataset are not expected to have \textit{complete} submask-aligned coverage of all elements one could recognize in or discuss about an image.\looseness-1

%-------------------------------------------------------------------------
\subsection{Collection Process}
\label{sec:collection}

We use Mephisto~\citep{urbanek2023mephisto} to host our task, pay crowdworkers to provide annotations on the dataset, and additionally run qualification steps. Workers that pass our qualifications are eligible to work on the main task which contains 3 stages:
\begin{enumerate}
  \item Workers are provided with the whole image, and asked to provide a short description of it. This is considered the \emph{standard caption}.
  \item Workers are provided with submasks of the image, one at a time starting with the leaves of the mask tree, displaying a SAM-selected region of the image as well as an indicator for where that region comes from. They are generally asked to provide a label and complete description for the pictured region, though are allowed to mark the region as `uninteresting' and only provide a label, or `bad' and provide nothing. These options allow us to focus worker time on useful annotations and help capture some of the noise of the automatic selection process. This is shown in Figure~\ref{fig:mask-procedure}. For masks that contain submasks, workers are also provided with overlays that show the regions already annotated, and are asked to annotate in terms of what has already been written.
  \item After completing all the submasks, the worker is then shown the complete image again and asked to provide an overall description, paying attention to the relationship between previously annotated regions.
\end{enumerate}
An in-depth description of the filtering and quality assurance process can be found in Appendix~\ref{sec:annotating} while the Datasheet~\citep{gebru2021datasheets} is available in Appendix \ref{sec:datasheet}. Complete annotation instructions, dataset download links as well as reproducible code are avaliable on our GitHub\footnote{\url{https://github.com/facebookresearch/DCI}}. The DCI dataset is released under the CC-BY-NC license. %\footnote{Lin}.

%-------------------------------------------------------------------------
\subsection{Fitting DCI into 77 CLIP tokens}
\label{sec:dci-fit}

Ultimately, we collected an average of 1111 words (1279 CLIP tokens) per image, with a median of 941 words. This proves problematic for evaluating or fine-tuning CLIP-based VLMs given their maximum text token length of 77. Embedding pooling methods~\citep{chen2018enhancing} to extend the effective input size for text modeling is an active research area~\citep{chalkidis2022exploration, zhang2022poolingformer}, and current work suggests average-pooling embeddings over these longer descriptions would be ineffective.\looseness-1

One possible approach would be to utilize the subsections of the image while providing the corresponding subcaption, in a manner akin to a multi-modal multi-crop approach~\citep{caron2021unsupervised}. Still, even when considering just the 91,424 submasks, the average token length is nearly 200 per caption. 
We instead use the longer context capabilities of Llama2~\citep{touvron2023llama} to summarize down the overall information in the image into CLIP-consumable portions. We generate multiple captions for each image and submask, using prompts that attempt to summarize down recursively until the result is in bounds. As this modification to the dataset is generated automatically, the summarizations may have introduced noise, and may not capture all of the detail in the full original captions. Summarizations also occasionally mix references or include context in a submask that isn't the main focus. Still, the summaries are fairly high quality and more dense than those found in other datasets, especially when using more than one distinct summarization per image. We also prompt the LLM to generate negatives from these summaries, achieving a set of particularly hard negatives for CLIP to evaluate. We call this version of the dataset \textbf{summarized DCI (sDCI)}. Examples of full caption, \emph{LLM summary} and \emph{LLM negative} are included in Figure~\ref{fig:dac_vs_dci} and contrasted with DAC~\citep{doveh2023dense} data. More detail including the prompts used can be found in Appendix \ref{sec:summarization}. 

Ultimately, this fitting step produces a \textit{lower bound} on the level of vision-language understanding `resolution' that the overall DCI dataset is capable of evaluating a model for. As newer models arise that are able to handle embedding much larger quantities of text content, it will be possible to make full use of DCI's original annotated captions.

%-------------------------------------------------------------------------
\subsection{Statistics}

\begin{table}[ht!]
\centering
\setlength{\tabcolsep}{4pt}
\begin{tabular}{rl|rrrr}
\multicolumn{2}{c|}{Dataset} & \multicolumn{1}{c}{Imgs} & \multicolumn{1}{c}{Caps} & \multicolumn{1}{c}{Toks/Cap} & \multicolumn{1}{c}{Toks/Img} \\ \midrule
\multicolumn{2}{l|}{DCI}  & 7,805 & 7,805 & 1,282.09 & \textbf{1,282.09} \\ 
\multicolumn{2}{l|}{$\textrm{DCI}_{sub}$} & 96,007 & 96,007 & 199.33 & 199.33 \\ 
\multicolumn{2}{l|}{sDCI} & 7,805 & 87,268 & 49.21 & 536.00 \\ 
\multicolumn{2}{l|}{$\textrm{sDCI}_{sub}$} & 96,007 & 714,630 & 36.60 & 263.01 \\ \midrule
 \multicolumn{2}{l|}{$\textrm{LN}_{COCO}$} & 142,845 & 142,845 & 49.11 & \textbf{49.11} \\
 \multicolumn{2}{l|}{$\textrm{LN}_{COCO<77}$} & 127,456 & 127,456 & 43.70 & 43.70 \\ \hline
 \multicolumn{2}{l|}{$\textrm{COCO}$} & 123,287 & 616,767 & 13.54 & 67.74 \\ \bottomrule
\end{tabular}
\caption{Comparison of DCI dataset statistics to other datasets, focusing on average CLIP tokens per image or caption. Note the \textbf{26x} difference between DCI and the previous longest annotated dataset, Localized Narratives (LN). $sub$ denotes including submasks and their descriptions as examples, and sDCI refers to the LLM-summarized version of DCI that fits captions to 77 tokens (Sec. \ref{sec:dci-fit}), while $\textrm{LN}_{COCO<77}$ simply drops examples longer than 77 tokens ($\sim10.8\%$).}
\label{tab:dataset-statistics}
\end{table}

All-in-all the Densely Captioned Images dataset is far more dense than Localized Narratives on COCO images~\citep{ponttuset2020connecting} (later referred to as $LN_{COCO}$) and nearly $100\times$ more dense than standard COCO captions~\citep{chen2015microsoft}. After reducing to CLIP-bounded summaries, it still contains more text density than both. Complete details can be found in Table~\ref{tab:dataset-statistics}. 

Here we see that the multiple-summarization method of sDCI produces fairly similar token per image values to the original dataset while keeping individual captions' token lengths in bounds for CLIP. To get Localized Narratives into the 77 token bound, we simply drop longer examples.

% \section{Methodology}
\section{Evaluating VLMs with summarized DCI}
\label{sec:methods}

%-------------------------------------------------------------------------
\subsection{Methodology}

\begin{table*}[ht]
\centering
\begin{tabular}{llll|rrrrrr}
\multicolumn{4}{c|}{}  & \multicolumn{2}{c}{\textbf{All}}                               & \multicolumn{2}{c}{\textbf{All Pick5}}                         & \multicolumn{1}{c}{\textbf{Base}} & \multicolumn{1}{c}{\textbf{All}}       \\
\multicolumn{4}{c|}{\textbf{Model}}  & \multicolumn{1}{c}{SCM}  & \multicolumn{1}{c}{Neg}   & \multicolumn{1}{c}{SCM}  & \multicolumn{1}{c}{Neg}   & \multicolumn{1}{c}{Neg}  & \multicolumn{1}{c}{Hard Negs} \\ \midrule
\multicolumn{4}{c|}{CLIP Baseline~\citep{CLIP}}& 40.06\% & 60.79\% & 11.21\% & 24.06\% & 67.56\% & 41.34\% \\
\multicolumn{4}{c|}{NegCLIP~\citep{aroNegClip}} & \textbf{43.35\%} & 56.00\% &	\textbf{13.22\%} &	4.82\% &	76.69\% &	50.84\% \\
\multicolumn{4}{c|}{BLIP~\citep{BLIP}}& 39.13\% &	54.02\% &	10.73\% &	5.51\% &	63.41\% &	53.23\% \\
\multicolumn{4}{c|}{Flava~\citep{singh2022flava}}& 38.08\% &	47.99\% &	8.01\% &	9.82\% &	11.6\% &	45.59\% \\
\multicolumn{4}{c|}{X-VLM~\citep{xvlm}}& 38.45\% &	53.46\% &	10.96\% &	5.10\% &	44.29\% &	52.42\% \\
\multicolumn{4}{c|}{$\textrm{DAC}_{LLM}$\citep{doveh2023dense}} & 37.45\% & 81.71\% &  8.13\% & 37.84\% &  \textbf{90.56\%} & 71.21\% \\
\multicolumn{4}{c|}{$\textrm{DAC}_{SAM}$~\citep{doveh2023dense}} &   37.90\% & \textbf{84.17\%} & 6.70\% & \textbf{39.94\%} & 89.66\% &  \textbf{73.61\%}\\
\end{tabular}
\caption{sDCI test result: We compare existing baselines on our Subcrop-Caption Matching (SCM) and negatives tests. Additional results are available in Table~\ref{tab:dci-full-test} in the Appendix. We note our best model fine-tuned on sDCI from section \ref{sec:dci-fine-tuning} achieved 64.02\% and 31.60\% on a held-out test of \textbf{All SCM} and \textbf{All SCM Pick5} respectively, setting an upper bound for model performance.}
\label{tab:dci-test}
\end{table*}

Using the 7805 images in the summarized Densely Captioned Images (sDCI) dataset, we construct a few different evaluations. As noted above, the ability to select multiple submasks from the same image and include them in the same batch allows us to create a CLIP-style test, wherein the model can evaluate a full batch of images and captions and score correctly which caption belongs to which image. As we provide models with a crop around the selected masks, we call this \emph{Subcrop-Caption Matching (SCM)}, and we use a batch size of 8. We can run against our LLM-generated negatives as well. Given that LLM-summarization has provided us with multiple captions and negatives per image and submask, we supply the first unless noted otherwise. With this in mind, we construct 6 evaluations as follows: \\
\textbf{[All SCM]:} Group each image with their subcrops, alongside one  summarized caption per subcrop. Then use the model to find the most likely caption associated to each subcrop. This test measures the ability of the VLM to distinguish between the different parts that compose an image.\footnote{Since we used sDCI to fit current models token length, it is possible that some of the summaries remove the information that make possible to distinguish between the captions. Ideally this test should be performed on the non-summarized version once VLMs can handle 1000+ tokens.} \\
\textbf{[All Neg]:} Select one LLM summarized caption and the corresponding LLM-generated negative for each image and subcrop. Score a model on its ability to distinguish between the positive and negative.\\
\textbf{[All Pick5-SCM]:} Use the same setup as \textbf{All SCM}, but rather than using only one caption per subcrop, we use 5 LLM generated captions per subcrop. We score a model as succeeding only when the worst-scoring positive caption scores higher than the best-scoring caption of any other image in the batch. This test evaluates if the representation space is structured such that captions belonging to a specific image are closest to the target image in the space. \\
\textbf{[All Pick5-Neg]:} Use the same setup as \textbf{All Neg}, but rather than using one caption, we use 5 LLM summarized captions for each image and subcrop. If any of these captions score worse than the negative, the model fails the example.\\
\textbf{[Base Neg]:} Using only the 7805 base images without subcrops, evaluate the model's ability to distinguish between an LLM generated caption and its corresponding LLM-generated negative. Note, this is a strict subset of \textbf{All Neg}, though these captions are on the longer side on average and cover a different distribution.\\
\textbf{[All Hard-Negs]:} Using the same setup as \textbf{All Neg}, but rather than using a single negative, use the negative across all LLM-generated negatives that CLIP scores highest.

\subsection{Results}
\label{sec:eval-results}

We compare in Table \ref{tab:dci-test} the sDCI performances given by different state-of-the-art models: CLIP~\citep{CLIP}, NegCLIP~\citep{aroNegClip}, BLIP~\citep{BLIP}, Flava~\citep{singh2022flava} and X-VLM~\citep{xvlm}. Additional experiments on different architectures and pretraining datasets are available in Table \ref{tab:dci-full-test} (see Appendix).
The CLIP baseline starts at 40.12\% on \textbf{All SCM} and 60.63\% on \textbf{All Neg}. The only model to improve over CLIP on SCM tasks is NegCLIP, which follows the fact that the hard image negatives that NegCLIP is trained on provide the most similar task to what we test of any of these models. None of the models trained without an explicit CLIP-loss component outperform CLIP on SCM tasks, but DAC ultimately performs the worst.

Performance on the Pick5 variations of each task follow the trends of the standard performance. Performance on \textbf{Base Neg} for Flava point to a weakness in comparing longer text examples, given the significant drop from 47.99\% to 11.6\% that is not demonstrated in other models.

Interestingly, models trained absent of CLIP (BLIP, Flava, X-VLM) experience a far less noticeable drop in performance between \textbf{All Neg} and \textbf{All Hard Negs}. This validates that sDCI's CLIP-hard negatives are not simply a higher proportion of `impossible' negatives, but rather capture some underlying trait about the negatives that CLIP models and their descendants all struggle with.

None of the models presented perform well across all of the sDCI test set. Given each of the CLIP-style models have some kind of advantage on this test set due to being trained on some objective that sDCI directly evaluates, we expect that the BLIP, Flava, and X-VLM scores are somewhat representative for existing state-of-the-art models' true performance on this test set.

\section{Using summarized DCI as fine-tuning dataset}
\label{sec:dci-fine-tuning}

\begin{table*}[ht!]
\centering
\setlength{\tabcolsep}{6pt}
\begin{tabular}{l|rrrr|rrr}
&  \multicolumn{4}{c|}{\textbf{ARO}} & \multicolumn{3}{c}{\textbf{VL-Checklist}} \\
\multicolumn{1}{c|}{\textbf{Model}} & \multicolumn{1}{c}{VG-R} & \multicolumn{1}{c}{VG-A} & \multicolumn{1}{c}{COCO} & \multicolumn{1}{c|}{FLICKR} & \multicolumn{1}{c}{Object} & \multicolumn{1}{c}{Attribute} & \multicolumn{1}{c}{Relation} \\ \midrule
$\textrm{sDCI}_{P1}$  & 76.23\% & 67.56\% & 88.58\% & 91.30\% & 80.71\% & 68.69\% & 70.12\% \\
$\textrm{sDCI}_{P1NL0}$  & 57.34\% & 61.98\% & 39.36\% & 44.62\% & \textit{88.37}\% & 70.42\% & 61.28\% \\ \midrule
% $\textrm{DAC}_{LLM_{10000}}$  & 61.53$\pm$0.05\% & 63.89$\pm$0.02\% & 46.28$\pm$1.5\% & 59.41$\pm$1.9\% & 66.90$\pm$0.01\% & 57.4$\pm$0.13\% & 56.96$\pm$0.07\% \\
$\textrm{DAC}_{LLM_{10,000}}$  & 61.53\% & 63.89\% & 46.28$\pm$1.5\% & 59.41$\pm$1.9\% & 66.90\% & 57.4\% & 56.96\% \\
$\textrm{DAC}_{LLM_{100,000}}$  & 61.0\% & 63.6\% & 48.2\% & 61.42\% & 66.87\% & 57.22\% & 57.18\% \\
$\textrm{DAC}_{LLM_{500,000}}$  & 60.1\% & 63.8\% & 50.2\% & 61.6\% & 66.54\% & 57.39\% & 56.77\% \\
%\hline
$\textrm{DAC}_{LLM_{3,000,000}}$ & \textbf{81.28}\% & \textbf{73.91\%} & \textbf{94.47\%} & \textbf{95.68\%} & 87.30\% & \textbf{77.27\%} & 86.41\% \\
$\textrm{DAC}_{SAM_{3,000,000}}$ & 77.16\% & 70.5\% & 91.22\% & 93.88\% & \textbf{88.50\%} & 75.83\% & \textbf{89.75\%} \\
\midrule \midrule
CLIP Baseline~\cite{CLIP} & 59.98\% & 63.18\% & 47.9\% & 60.2\% & 81.17\% & 67.67\% & 61.95\% \\
BLIP2~\cite{BLIP2} & 41.16\% & 71.25\% & 13.57\% & 13.72\% & 84.14\% & 80.12\% & 70.72\% \\
NegCLIP~\cite{aroNegClip} & 81\% & 71\% & 86\% & 91\% & 81.35\% & 72.24\% & 63.53\% \\
SVLC~\cite{doveh2023teaching} & 80.61\% & 73.03\% & 84.73\% & 91.7\% & 85\% & 71.97\% & 68.95\% \\
\end{tabular}
\caption{$\textrm{sDCI}$ fine-tuned CLIP performance against the ARO and VL-Checklist benchmark. We compare CLIP fine-tuned with sDCI against models fine-tuned using DAC captions. Since the DAC dataset contains 3M images whereas sDCI contains only 7805 images, we performed an ablation of the number of training images used in the DAC dataset. In this instance, $\textrm{DAC}_{LLM_{10000}}$ refer to fine-tuning CLIP using only 10,000 images from DAC. We plot the mean across 5 different seeds and display the standard deviation when it is above 1\% accuracy. We observe that training on sDCI lead to significant improvement in comparison to DAC for a comparable number of examples.}
\label{tab:standard-benchmarks}
\end{table*}

To evaluate the use and difficulty of the sDCI dataset for training, we fine-tune state-of-the-art models with it. In particular, we use a ViT/32B CLIP model in all of our experiments, requiring use of the CLIP-bounded version of our dataset. We split sDCI into 7800 train, 100 validation, 112 test samples for this purpose. We use a training batch size of 32 and a learning rate of \texttt{5e-5} for all experiments, and run for up to 10 epochs. We train using both standard CLIP loss as well as an additional Negatives loss component, which follows the `text negative' of NegCLIP~\citep{aroNegClip}. Given the tiny size of our finetuning sets relative to the 400M pretraining images, we use LoRA~\citep{hu2021lora} to reduce the trainable parameters. We train a model with and without negatives loss.

In order to make good use of the multiple summarized captions we have per image and submask, we randomly select one to be used in each individual epoch. We call this method \emph{Pick1}. We describe this method and other ablations we attempted in more detail in Appendix~\ref{sec:more-ablation}.

We follow the experimental setup of DAC~\citep{doveh2023dense} by evaluating our sDCI fine-tuned CLIP on the ARO and VL-Checklist benchmarks.
%We compare our fine-tuned sDCI models with DAC~\citep{doveh2023dense} on the ARO and VL-Checklist benchmark. 
We compare to DAC directly as it is the most similar work to ours in attempting to increase caption density. As noted in Figure~\ref{fig:dac_vs_dci}, these automatically generated captions are generally noisy. As DAC is using 3M images for fine-tuning, we performed a small ablation on the number of DAC images to use for fine-tuning to be similar to our base image count (10,000 compared to our 8,012), or to our full mask count (100,000 compared to our 99,445).\looseness-1 %For the smaller ablation, we average over 5 seeds to help remedy sample size concerns.

\subsection{Results}

In Table \ref{tab:standard-benchmarks}, we show that, while the DCI Pick1 model trained with negatives loss ($\textrm{sDCI}_{P1}$) does not reach the performance of DAC models trained on 3M images, it does improve over the CLIP baseline on most metrics\footnote{The decreased performance on VL-Object may be explained by our LLM-generated negatives not closely covering the test set negatives.}, and outperforms some baselines trained on more data. $\textrm{sDCI}_{P1}$ does however outperform both sample-limited ablations of DAC, suggesting that a small number of highly aligned image to dense text pairs are more effective for training models than larger quantities of more loosely aligned or sparse data. 
Unsurprisingly, the version trained without negatives loss, $\textrm{sDCI}_{P1NL0}$, does not improve across most benchmarks, and even somewhat degrades when compared to the CLIP baseline.\footnote{The degradation is likely due to the distribution shift and small sample size, given the training objective is the same as CLIP.} Of note however is the significant bump in VL-Object, alongside some improvement to VL-Attribute.
Improvements here suggest that the sDCI dataset successfully includes more object, and to a lesser degree attribute, information than the captions in the source dataset for CLIP. It does, however, point to a limitation of using the LLM summarizations and not incorporating mask information, as relational information is sometimes lost.

%-------------------------------------------------------------------------

\section{Conclusion and Future Work}
\label{sec:conclusion}

We introduce the Densely Captioned Images dataset, and display clear use for it as a evaluation benchmark. We also show initial potential for using the dataset for fine-tuning.
Given that in order to evaluate today's models on DCI we had to reduce the size of the text to only 77 tokens, DCI should prove to be useful for a longer period of time as models that are able to consume and utilize larger amounts of text context become the norm. We envision that in those cases the full human annotated captions without length reduction would be provided.
Today's context size limitation also prevented us from fine-tuning existing models on the highly aligned text-image data within DCI, as existing models don't have enough context size to handle the full text, but the dataset isn't nearly large enough to pre-train a new set of models that could use the full text. It could be relevant to treat developing highly aligned text-image datasets in a similar manner to that used in machine translation for low-resource languages, which run into a similar issue with cost and difficulty to collect. This area of work has relied on automated methods such as bitext mining~\citep{nllbteam2022language} to bootstrap up from an initial set of expertly collected examples, which DCI may already provide the foundation for.
Further, we haven't attempted to incorporate the pixel-level masks that the dataset has in any of our experiments, instead opting to use crops around the masks to retain parity with our test set. This dataset is unique for both the extreme density and high degree of alignment present, and in this introductory work we've only scratched the surface of using this information to its fullest extent.

{
    \small
    \bibliographystyle{plainnat}
    \bibliography{main}

\begin{thebibliography}{45}
\providecommand{\natexlab}[1]{#1}
\providecommand{\url}[1]{\texttt{#1}}
\expandafter\ifx\csname urlstyle\endcsname\relax
  \providecommand{\doi}[1]{doi: #1}\else
  \providecommand{\doi}{doi: \begingroup \urlstyle{rm}\Url}\fi

\bibitem[Abbas et~al.(2023)Abbas, Tirumala, Simig, Ganguli, and Morcos]{abbas2023semdedup}
Amro Kamal~Mohamed Abbas, Kushal Tirumala, Daniel Simig, Surya Ganguli, and Ari~S. Morcos.
\newblock Semdedup: Data-efficient learning at web-scale through semantic deduplication.
\newblock In \emph{ICLR 2023 Workshop on Mathematical and Empirical Understanding of Foundation Models}, 2023.
\newblock \url{https://openreview.net/forum?id=4vlGm9gv6c}.

\bibitem[Alayrac et~al.(2022)Alayrac, Donahue, Luc, Miech, Barr, Hasson, Lenc, Mensch, Millican, Reynolds, Ring, Rutherford, Cabi, Han, Gong, Samangooei, Monteiro, Menick, Borgeaud, Brock, Nematzadeh, Sharifzadeh, Binkowski, Barreira, Vinyals, Zisserman, and Simonyan]{alayrac2022flamingo}
Jean-Baptiste Alayrac, Jeff Donahue, Pauline Luc, Antoine Miech, Iain Barr, Yana Hasson, Karel Lenc, Arthur Mensch, Katie Millican, Malcolm Reynolds, Roman Ring, Eliza Rutherford, Serkan Cabi, Tengda Han, Zhitao Gong, Sina Samangooei, Marianne Monteiro, Jacob Menick, Sebastian Borgeaud, Andrew Brock, Aida Nematzadeh, Sahand Sharifzadeh, Mikolaj Binkowski, Ricardo Barreira, Oriol Vinyals, Andrew Zisserman, and Karen Simonyan.
\newblock Flamingo: a visual language model for few-shot learning, 2022.

\bibitem[Bordes et~al.(2023)Bordes, Shekhar, Ibrahim, Bouchacourt, Vincent, and Morcos]{bordes2023pug}
Florian Bordes, Shashank Shekhar, Mark Ibrahim, Diane Bouchacourt, Pascal Vincent, and Ari~S. Morcos.
\newblock Pug: Photorealistic and semantically controllable synthetic data for representation learning.
\newblock In \emph{Advances in Neural Information Processing Systems}, 2023.

\bibitem[Canny(1986)]{cannyfilter}
John Canny.
\newblock A computational approach to edge detection.
\newblock \emph{IEEE Transactions on Pattern Analysis and Machine Intelligence}, PAMI-8\penalty0 (6):\penalty0 679--698, 1986.
\newblock \doi{10.1109/TPAMI.1986.4767851}.

\bibitem[Caron et~al.(2021)Caron, Misra, Mairal, Goyal, Bojanowski, and Joulin]{caron2021unsupervised}
Mathilde Caron, Ishan Misra, Julien Mairal, Priya Goyal, Piotr Bojanowski, and Armand Joulin.
\newblock Unsupervised learning of visual features by contrasting cluster assignments, 2021.

\bibitem[Chalkidis et~al.(2022)Chalkidis, Dai, Fergadiotis, Malakasiotis, and Elliott]{chalkidis2022exploration}
Ilias Chalkidis, Xiang Dai, Manos Fergadiotis, Prodromos Malakasiotis, and Desmond Elliott.
\newblock An exploration of hierarchical attention transformers for efficient long document classification, 2022.

\bibitem[Chen et~al.(2018)Chen, Ling, and Zhu]{chen2018enhancing}
Qian Chen, Zhen-Hua Ling, and Xiaodan Zhu.
\newblock Enhancing sentence embedding with generalized pooling, 2018.

\bibitem[Chen et~al.(2015)Chen, Fang, Lin, Vedantam, Gupta, Dollar, and Zitnick]{chen2015microsoft}
Xinlei Chen, Hao Fang, Tsung-Yi Lin, Ramakrishna Vedantam, Saurabh Gupta, Piotr Dollar, and C.~Lawrence Zitnick.
\newblock Microsoft coco captions: Data collection and evaluation server, 2015.

\bibitem[Desai et~al.(2021)Desai, Kaul, Aysola, and Johnson]{desai2021redcaps}
Karan Desai, Gaurav Kaul, Zubin Aysola, and Justin Johnson.
\newblock Redcaps: web-curated image-text data created by the people, for the people, 2021.

\bibitem[Doveh et~al.(2023{\natexlab{a}})Doveh, Arbelle, Harary, Herzig, Kim, Cascante-bonilla, Alfassy, Panda, Giryes, Feris, Ullman, and Karlinsky]{doveh2023dense}
Sivan Doveh, Assaf Arbelle, Sivan Harary, Roei Herzig, Donghyun Kim, Paola Cascante-bonilla, Amit Alfassy, Rameswar Panda, Raja Giryes, Rogerio Feris, Shimon Ullman, and Leonid Karlinsky.
\newblock Dense and aligned captions (dac) promote compositional reasoning in vl models, 2023{\natexlab{a}}.

\bibitem[Doveh et~al.(2023{\natexlab{b}})Doveh, Arbelle, Harary, Panda, Herzig, Schwartz, Kim, Giryes, Feris, Ullman, and Karlinsky]{doveh2023teaching}
Sivan Doveh, Assaf Arbelle, Sivan Harary, Rameswar Panda, Roei Herzig, Eli Schwartz, Donghyun Kim, Raja Giryes, Rogerio Feris, Shimon Ullman, and Leonid Karlinsky.
\newblock Teaching structured vision\&language concepts to vision\&language models, 2023{\natexlab{b}}.

\bibitem[Fini et~al.(2023)Fini, Astolfi, Romero-Soriano, Verbeek, and Drozdzal]{fini2023improved}
Enrico Fini, Pietro Astolfi, Adriana Romero-Soriano, Jakob Verbeek, and Michal Drozdzal.
\newblock Improved baselines for vision-language pre-training.
\newblock \emph{Transactions on Machine Learning Research (TMLR)}, 2023.

\bibitem[Gebru et~al.(2021)Gebru, Morgenstern, Vecchione, Vaughan, Wallach, au2, and Crawford]{gebru2021datasheets}
Timnit Gebru, Jamie Morgenstern, Briana Vecchione, Jennifer~Wortman Vaughan, Hanna Wallach, Hal Daumé~III au2, and Kate Crawford.
\newblock Datasheets for datasets, 2021.

\bibitem[Hu et~al.(2021)Hu, Shen, Wallis, Allen-Zhu, Li, Wang, Wang, and Chen]{hu2021lora}
Edward~J. Hu, Yelong Shen, Phillip Wallis, Zeyuan Allen-Zhu, Yuanzhi Li, Shean Wang, Lu~Wang, and Weizhu Chen.
\newblock Lora: Low-rank adaptation of large language models, 2021.

\bibitem[Kirillov et~al.(2023)Kirillov, Mintun, Ravi, Mao, Rolland, Gustafson, Xiao, Whitehead, Berg, Lo, Dollár, and Girshick]{kirillov2023segment}
Alexander Kirillov, Eric Mintun, Nikhila Ravi, Hanzi Mao, Chloe Rolland, Laura Gustafson, Tete Xiao, Spencer Whitehead, Alexander~C. Berg, Wan-Yen Lo, Piotr Dollár, and Ross Girshick.
\newblock Segment anything, 2023.

\bibitem[Krishna et~al.(2016)Krishna, Zhu, Groth, Johnson, Hata, Kravitz, Chen, Kalantidis, Li, Shamma, Bernstein, and Li]{krishna2016visual}
Ranjay Krishna, Yuke Zhu, Oliver Groth, Justin Johnson, Kenji Hata, Joshua Kravitz, Stephanie Chen, Yannis Kalantidis, Li-Jia Li, David~A. Shamma, Michael~S. Bernstein, and Fei-Fei Li.
\newblock Visual genome: Connecting language and vision using crowdsourced dense image annotations, 2016.

\bibitem[Li et~al.(2022{\natexlab{a}})Li, Liu, Li, Zhang, Aneja, Yang, Jin, Hu, Liu, Lee, and Gao]{li2022elevater}
Chunyuan Li, Haotian Liu, Liunian~Harold Li, Pengchuan Zhang, Jyoti Aneja, Jianwei Yang, Ping Jin, Houdong Hu, Zicheng Liu, Yong~Jae Lee, and Jianfeng Gao.
\newblock Elevater: A benchmark and toolkit for evaluating language-augmented visual models, 2022{\natexlab{a}}.

\bibitem[Li et~al.(2022{\natexlab{b}})Li, Li, Xiong, and Hoi]{BLIP}
Junnan Li, Dongxu Li, Caiming Xiong, and Steven C.~H. Hoi.
\newblock {BLIP:} bootstrapping language-image pre-training for unified vision-language understanding and generation.
\newblock \emph{CoRR}, abs/2201.12086, 2022{\natexlab{b}}.

\bibitem[Li et~al.(2023{\natexlab{a}})Li, Li, Savarese, and Hoi]{BLIP2}
Junnan Li, Dongxu Li, Silvio Savarese, and Steven Hoi.
\newblock Blip-2: Bootstrapping language-image pre-training with frozen image encoders and large language models, 2023{\natexlab{a}}.

\bibitem[Li et~al.(2023{\natexlab{b}})Li, Wang, and Xie]{li2023inverse}
Xianhang Li, Zeyu Wang, and Cihang Xie.
\newblock An inverse scaling law for clip training, 2023{\natexlab{b}}.

\bibitem[Li et~al.(2023{\natexlab{c}})Li, Fan, Hu, Feichtenhofer, and He]{li2023scaling}
Yanghao Li, Haoqi Fan, Ronghang Hu, Christoph Feichtenhofer, and Kaiming He.
\newblock Scaling language-image pre-training via masking, 2023{\natexlab{c}}.

\bibitem[Lin et~al.(2023)Lin, Chen, Pathak, Zhang, and Ramanan]{lin2023visualgptscore}
Zhiqiu Lin, Xinyue Chen, Deepak Pathak, Pengchuan Zhang, and Deva Ramanan.
\newblock Visualgptscore: Visio-linguistic reasoning with multimodal generative pre-training scores, 2023.

\bibitem[Ma et~al.(2023)Ma, Hong, Gul, Gandhi, Gao, and Krishna]{crepe}
Zixian Ma, Jerry Hong, Mustafa~Omer Gul, Mona Gandhi, Irena Gao, and Ranjay Krishna.
\newblock Crepe: Can vision-language foundation models reason compositionally?, 2023.

\bibitem[Ordonez et~al.(2011)Ordonez, Kulkarni, and Berg]{NIPS2011_5dd9db5e}
Vicente Ordonez, Girish Kulkarni, and Tamara Berg.
\newblock Im2text: Describing images using 1 million captioned photographs.
\newblock In J.~Shawe-Taylor, R.~Zemel, P.~Bartlett, F.~Pereira, and K.Q. Weinberger, editors, \emph{Advances in Neural Information Processing Systems}, volume~24. Curran Associates, Inc., 2011.
\newblock \url{https://proceedings.neurips.cc/paper_files/paper/2011/file/5dd9db5e033da9c6fb5ba83c7a7ebea9-Paper.pdf}.

\bibitem[Pont-Tuset et~al.(2020)Pont-Tuset, Uijlings, Changpinyo, Soricut, and Ferrari]{ponttuset2020connecting}
Jordi Pont-Tuset, Jasper Uijlings, Soravit Changpinyo, Radu Soricut, and Vittorio Ferrari.
\newblock Connecting vision and language with localized narratives, 2020.

\bibitem[Radenovic et~al.(2023)Radenovic, Dubey, Kadian, Mihaylov, Vandenhende, Patel, Wen, Ramanathan, and Mahajan]{radenovic2023filtering}
Filip Radenovic, Abhimanyu Dubey, Abhishek Kadian, Todor Mihaylov, Simon Vandenhende, Yash Patel, Yi~Wen, Vignesh Ramanathan, and Dhruv Mahajan.
\newblock Filtering, distillation, and hard negatives for vision-language pre-training.
\newblock In \emph{Proceedings of the IEEE/CVF Conference on Computer Vision and Pattern Recognition}, pages 6967--6977, 2023.

\bibitem[Radford et~al.(2021)Radford, Kim, Hallacy, Ramesh, Goh, Agarwal, Sastry, Askell, Mishkin, Clark, Krueger, and Sutskever]{CLIP}
Alec Radford, Jong~Wook Kim, Chris Hallacy, Aditya Ramesh, Gabriel Goh, Sandhini Agarwal, Girish Sastry, Amanda Askell, Pamela Mishkin, Jack Clark, Gretchen Krueger, and Ilya Sutskever.
\newblock Learning transferable visual models from natural language supervision.
\newblock \emph{CoRR}, abs/2103.00020, 2021.

\bibitem[Schuhmann et~al.(2021)Schuhmann, Vencu, Beaumont, Kaczmarczyk, Mullis, Katta, Coombes, Jitsev, and Komatsuzaki]{schuhmann2021laion400m}
Christoph Schuhmann, Richard Vencu, Romain Beaumont, Robert Kaczmarczyk, Clayton Mullis, Aarush Katta, Theo Coombes, Jenia Jitsev, and Aran Komatsuzaki.
\newblock Laion-400m: Open dataset of clip-filtered 400 million image-text pairs, 2021.

\bibitem[Schuhmann et~al.(2022)Schuhmann, Beaumont, Vencu, Gordon, Wightman, Cherti, Coombes, Katta, Mullis, Wortsman, Schramowski, Kundurthy, Crowson, Schmidt, Kaczmarczyk, and Jitsev]{schuhmann2022laion5b}
Christoph Schuhmann, Romain Beaumont, Richard Vencu, Cade Gordon, Ross Wightman, Mehdi Cherti, Theo Coombes, Aarush Katta, Clayton Mullis, Mitchell Wortsman, Patrick Schramowski, Srivatsa Kundurthy, Katherine Crowson, Ludwig Schmidt, Robert Kaczmarczyk, and Jenia Jitsev.
\newblock Laion-5b: An open large-scale dataset for training next generation image-text models, 2022.

\bibitem[Sharma et~al.(2018)Sharma, Ding, Goodman, and Soricut]{sharma-etal-2018-conceptual}
Piyush Sharma, Nan Ding, Sebastian Goodman, and Radu Soricut.
\newblock Conceptual captions: A cleaned, hypernymed, image alt-text dataset for automatic image captioning.
\newblock In \emph{Proceedings of the 56th Annual Meeting of the Association for Computational Linguistics (Volume 1: Long Papers)}, pages 2556--2565, Melbourne, Australia, July 2018. Association for Computational Linguistics.
\newblock \doi{10.18653/v1/P18-1238}.
\newblock \url{https://aclanthology.org/P18-1238}.

\bibitem[Singh et~al.(2022)Singh, Hu, Goswami, Couairon, Galuba, Rohrbach, and Kiela]{singh2022flava}
Amanpreet Singh, Ronghang Hu, Vedanuj Goswami, Guillaume Couairon, Wojciech Galuba, Marcus Rohrbach, and Douwe Kiela.
\newblock {FLAVA:} {A} foundational language and vision alignment model.
\newblock In \emph{CVPR}, 2022.

\bibitem[Srinivasan et~al.(2021)Srinivasan, Raman, Chen, Bendersky, and Najork]{wit}
Krishna Srinivasan, Karthik Raman, Jiecao Chen, Mike Bendersky, and Marc Najork.
\newblock Wit: Wikipedia-based image text dataset for multimodal multilingual machine learning.
\newblock In \emph{Proceedings of the 44th International ACM SIGIR Conference on Research and Development in Information Retrieval (SIGIR '21)}, 2021.
\newblock \url{https://arxiv.org/abs/2103.01913}.

\bibitem[Team et~al.(2022)Team, Costa-jussà, Cross, Çelebi, Elbayad, Heafield, Heffernan, Kalbassi, Lam, Licht, Maillard, Sun, Wang, Wenzek, Youngblood, Akula, Barrault, Gonzalez, Hansanti, Hoffman, Jarrett, Sadagopan, Rowe, Spruit, Tran, Andrews, Ayan, Bhosale, Edunov, Fan, Gao, Goswami, Guzmán, Koehn, Mourachko, Ropers, Saleem, Schwenk, and Wang]{nllbteam2022language}
NLLB Team, Marta~R. Costa-jussà, James Cross, Onur Çelebi, Maha Elbayad, Kenneth Heafield, Kevin Heffernan, Elahe Kalbassi, Janice Lam, Daniel Licht, Jean Maillard, Anna Sun, Skyler Wang, Guillaume Wenzek, Al~Youngblood, Bapi Akula, Loic Barrault, Gabriel~Mejia Gonzalez, Prangthip Hansanti, John Hoffman, Semarley Jarrett, Kaushik~Ram Sadagopan, Dirk Rowe, Shannon Spruit, Chau Tran, Pierre Andrews, Necip~Fazil Ayan, Shruti Bhosale, Sergey Edunov, Angela Fan, Cynthia Gao, Vedanuj Goswami, Francisco Guzmán, Philipp Koehn, Alexandre Mourachko, Christophe Ropers, Safiyyah Saleem, Holger Schwenk, and Jeff Wang.
\newblock No language left behind: Scaling human-centered machine translation, 2022.

\bibitem[Thomee et~al.(2016)Thomee, Shamma, Friedland, Elizalde, Ni, Poland, Borth, and Li]{Thomee_2016}
Bart Thomee, David~A. Shamma, Gerald Friedland, Benjamin Elizalde, Karl Ni, Douglas Poland, Damian Borth, and Li-Jia Li.
\newblock {YFCC}100m.
\newblock \emph{Communications of the {ACM}}, 59\penalty0 (2):\penalty0 64--73, jan 2016.
\newblock \doi{10.1145/2812802}.
\newblock \url{https://doi.org/10.1145%2F2812802}.

\bibitem[Thrush et~al.(2022)Thrush, Jiang, Bartolo, Singh, Williams, Kiela, and Ross]{winoground}
Tristan Thrush, Ryan Jiang, Max Bartolo, Amanpreet Singh, Adina Williams, Douwe Kiela, and Candace Ross.
\newblock Winoground: Probing vision and language models for visio-linguistic compositionality, 2022.

\bibitem[Touvron et~al.(2023)Touvron, Martin, Stone, Albert, Almahairi, Babaei, Bashlykov, Batra, Bhargava, Bhosale, Bikel, Blecher, Ferrer, Chen, Cucurull, Esiobu, Fernandes, Fu, Fu, Fuller, Gao, Goswami, Goyal, Hartshorn, Hosseini, Hou, Inan, Kardas, Kerkez, Khabsa, Kloumann, Korenev, Koura, Lachaux, Lavril, Lee, Liskovich, Lu, Mao, Martinet, Mihaylov, Mishra, Molybog, Nie, Poulton, Reizenstein, Rungta, Saladi, Schelten, Silva, Smith, Subramanian, Tan, Tang, Taylor, Williams, Kuan, Xu, Yan, Zarov, Zhang, Fan, Kambadur, Narang, Rodriguez, Stojnic, Edunov, and Scialom]{touvron2023llama}
Hugo Touvron, Louis Martin, Kevin Stone, Peter Albert, Amjad Almahairi, Yasmine Babaei, Nikolay Bashlykov, Soumya Batra, Prajjwal Bhargava, Shruti Bhosale, Dan Bikel, Lukas Blecher, Cristian~Canton Ferrer, Moya Chen, Guillem Cucurull, David Esiobu, Jude Fernandes, Jeremy Fu, Wenyin Fu, Brian Fuller, Cynthia Gao, Vedanuj Goswami, Naman Goyal, Anthony Hartshorn, Saghar Hosseini, Rui Hou, Hakan Inan, Marcin Kardas, Viktor Kerkez, Madian Khabsa, Isabel Kloumann, Artem Korenev, Punit~Singh Koura, Marie-Anne Lachaux, Thibaut Lavril, Jenya Lee, Diana Liskovich, Yinghai Lu, Yuning Mao, Xavier Martinet, Todor Mihaylov, Pushkar Mishra, Igor Molybog, Yixin Nie, Andrew Poulton, Jeremy Reizenstein, Rashi Rungta, Kalyan Saladi, Alan Schelten, Ruan Silva, Eric~Michael Smith, Ranjan Subramanian, Xiaoqing~Ellen Tan, Binh Tang, Ross Taylor, Adina Williams, Jian~Xiang Kuan, Puxin Xu, Zheng Yan, Iliyan Zarov, Yuchen Zhang, Angela Fan, Melanie Kambadur, Sharan Narang, Aurelien Rodriguez, Robert Stojnic, Sergey Edunov, and Thomas
  Scialom.
\newblock Llama 2: Open foundation and fine-tuned chat models, 2023.

\bibitem[Urbanek and Ringshia(2023)]{urbanek2023mephisto}
Jack Urbanek and Pratik Ringshia.
\newblock Mephisto: A framework for portable, reproducible, and iterative crowdsourcing, 2023.

\bibitem[Xu et~al.(2023{\natexlab{a}})Xu, Xie, Huang, Yu, Howes, Ghosh, Zettlemoyer, and Feichtenhofer]{xu2023cit}
Hu~Xu, Saining Xie, Po-Yao Huang, Licheng Yu, Russell Howes, Gargi Ghosh, Luke Zettlemoyer, and Christoph Feichtenhofer.
\newblock Cit: Curation in training for effective vision-language data.
\newblock \emph{arXiv preprint arXiv:2301.02241}, 2023{\natexlab{a}}.

\bibitem[Xu et~al.(2023{\natexlab{b}})Xu, Xie, Tan, Huang, Howes, Sharma, Li, Ghosh, Zettlemoyer, and Feichtenhofer]{xu2023demystifying}
Hu~Xu, Saining Xie, Xiaoqing~Ellen Tan, Po-Yao Huang, Russell Howes, Vasu Sharma, Shang-Wen Li, Gargi Ghosh, Luke Zettlemoyer, and Christoph Feichtenhofer.
\newblock Demystifying clip data, 2023{\natexlab{b}}.

\bibitem[Young et~al.(2014)Young, Lai, Hodosh, and Hockenmaier]{young-etal-2014-image}
Peter Young, Alice Lai, Micah Hodosh, and Julia Hockenmaier.
\newblock From image descriptions to visual denotations: New similarity metrics for semantic inference over event descriptions.
\newblock \emph{Transactions of the Association for Computational Linguistics}, 2:\penalty0 67--78, 2014.
\newblock \doi{10.1162/tacl_a_00166}.
\newblock \url{https://aclanthology.org/Q14-1006}.

\bibitem[Yu et~al.(2023)Yu, Shi, Pasunuru, Muller, Golovneva, Wang, Babu, Tang, Karrer, Sheynin, Ross, Polyak, Howes, Sharma, Xu, Tamoyan, Ashual, Singer, Li, Zhang, James, Ghosh, Taigman, Fazel-Zarandi, Celikyilmaz, Zettlemoyer, and Aghajanyan]{yu2023scaling}
Lili Yu, Bowen Shi, Ramakanth Pasunuru, Benjamin Muller, Olga Golovneva, Tianlu Wang, Arun Babu, Binh Tang, Brian Karrer, Shelly Sheynin, Candace Ross, Adam Polyak, Russell Howes, Vasu Sharma, Puxin Xu, Hovhannes Tamoyan, Oron Ashual, Uriel Singer, Shang-Wen Li, Susan Zhang, Richard James, Gargi Ghosh, Yaniv Taigman, Maryam Fazel-Zarandi, Asli Celikyilmaz, Luke Zettlemoyer, and Armen Aghajanyan.
\newblock Scaling autoregressive multi-modal models: Pretraining and instruction tuning, 2023.

\bibitem[Yuksekgonul et~al.(2023)Yuksekgonul, Bianchi, Kalluri, Jurafsky, and Zou]{aroNegClip}
Mert Yuksekgonul, Federico Bianchi, Pratyusha Kalluri, Dan Jurafsky, and James Zou.
\newblock When and why vision-language models behave like bags-of-words, and what to do about it?
\newblock In \emph{International Conference on Learning Representations}, 2023.
\newblock \url{https://openreview.net/forum?id=KRLUvxh8uaX}.

\bibitem[Zeng et~al.(2021)Zeng, Zhang, and Li]{xvlm}
Yan Zeng, Xinsong Zhang, and Hang Li.
\newblock Multi-grained vision language pre-training: Aligning texts with visual concepts.
\newblock \emph{arXiv preprint arXiv:2111.08276}, 2021.

\bibitem[Zhang et~al.(2022)Zhang, Gong, Shen, Li, Lv, Duan, and Chen]{zhang2022poolingformer}
Hang Zhang, Yeyun Gong, Yelong Shen, Weisheng Li, Jiancheng Lv, Nan Duan, and Weizhu Chen.
\newblock Poolingformer: Long document modeling with pooling attention, 2022.

\bibitem[Zhao et~al.(2023)Zhao, Zhang, Zhu, Shen, Lee, Lu, and Yin]{zhao2023vlchecklist}
Tiancheng Zhao, Tianqi Zhang, Mingwei Zhu, Haozhan Shen, Kyusong Lee, Xiaopeng Lu, and Jianwei Yin.
\newblock Vl-checklist: Evaluating pre-trained vision-language models with objects, attributes and relations, 2023.

\end{thebibliography}
}

\clearpage
\setcounter{page}{1}
%\maketitlesupplementary
\beginappendix

%----------------------------------------
\section{LLM Summarization and Negatives}
\label{sec:summarization}

In this section we discuss the prompting strategies we used on the LLaMA 70B-chat model to produce our summarized captions and negatives.

\subsection{Summarization}

In order to generate multiple summaries for each image and subsection of the image, we used the following method. First, we attempted to create a first-pass summary of either the full image or of masks with the following prompts: \\
\textbf{Full Image:} \texttt{You are given a full-text description of an image. You should summarize it into about 65 words, being sure to include as much salient visual information as possible given the 65 word constraint, especially information from the start of the original desc- ription. The new description should apply for the original image. Respond with only the summary, in one line.} \\

\textbf{Submask:} \texttt{You are given a description of part of an image. You should summarize it into a single line no longer than 65 words, being sure to include as much salient visual information as possible given the 65. Don't include any details not present in the provided description. Respond with only the summary, in one line.}

This provided us with once caption per image and submask, which is used universally as the \textit{first} caption for each. We, however, wanted to have more options to capture potential details that may have been missed by the first summarization. To that end we repeated the following prompt until we had at least 6 captions per.

\textbf{Multi-caption Prompt:} \texttt{You are providing descriptions of an image. The goal is to create a list of summarized descriptions that all accurately describe the same image, but may pay attention to slightly different details. A complete description will be provided. Complete 5 entries in this list, each a few sentences long. Provide in the format:\textbackslash{}n1. <description>\textbackslash{}n2. ...}

\subsection{Negatives}

To generate negatives, we came up with three distinct prompts to produce negatives when given a summary. All of the negatives are generated from the \textit{first} summarized caption. Each attempted to raise different kinds of reasoning, either for simple basic edits, changes to structural information, or reconstructing a new sentence from the same bag of words.\\
\textbf{basic:} \texttt{You are given the description for an image. You should provide a mostly similar description, changing the original one slightly, but introducing enough significant differences such that the two descriptions could not possibly be for the same image. Keep the description length the same. Provide just the updated description.}\\
\textbf{layout:} \texttt{You are given the description for an image. You should provide a mostly similar description, with one part of it changed in a way that visibly alters the structure, layout, or content of the image. The change should introduce enough difference such that the two descriptions could not be for the same image. Keep the description length the same. Provide just the updated description.}\\
\textbf{swaps:} \texttt{You are given a description. Selecting ONLY from the same words, construct a random response with the same words in a completely new order. The new description should not remain accurate to the first. Try to use all of the words from the original description, and keep the length the same, but use the words nearly randomly such that the scene makes less sense. Try to pair nouns and adjectives differently than in the original.}\\

Ultimately, in initial ablations we found that the \textbf{swaps} prompt produced the best model performance, so we use these negatives as the default.

\subsection{Common Issues}

One main issue with LLM-based summarization is that LLaMA2 was not particularly effective at producing answers within the specified word boundary. Thus, we additionally use prompts that attempt to get the model to shorten the output response so long as it is somewhat close in length. If after enough tries this didn't work, we'd return to the original prompt.

\textbf{Reduce: } \texttt{You are given the description for a scene. Summarize and rephrase the description such that the new description is a few words shorter, but retains key information.}

Another issue is that the chat fine-tuning on the LLaMA2 model made it frequently preface the response with additional affirmative text, like ``sure I can do that...''. Often this could be removed by simply removing the everything before the last line break.

\subsection{Output Text Examples}

Below we provide an example of a summarization for a base image (which includes details from all of the subimages during the prompting stage) as well as a summarization for one submask. We also share one example of each type of type of negative for both.

\subsubsection{Base Image}

\paragraph{Description:} The upper floor of an ornate design building with the words ``Dem Wahren, Schonen'' visible on its top. There's a light source with five light bulbs decorating each pillar, at least two marble statues and multiple arches between each pillar. An ornate building with at least two floors. There's four small pillars with three black guard rails and four light stands on the pillars. There's two marble statues on either end of the upper floor. The upper floor also has five arches with decorative symbols above them. There's six visible tall supporting pillars on either side of the arches. On the top is the visible words: Dem Wahren Schoenon. On the bottom floor is three arches with a black light hanging from each of them. The inner arches and interiors are somewhat visible. In front of the building is a black streetlight. ...(submask text omitted)...

\paragraph{Selected Summaries (3/8):} \textit{(1)} An ornate building with four pillars, three arches, and a decorative facade. The building has a upper floor with five light bulbs, two marble statues, and multiple arches. The lower floor has three black guard rails and a streetlight. The image includes various decorative elements such as swirling branches, leaves, and faces.\\
\textit{(2)} The image shows the upper floor of an ornate building with five pillars, each with a light source and decorative elements. The building has a grand entrance with multiple arches and marble statues. The words ``Dem Wahren, Schonen'' are visible on the top of the building.\\
\textit{(3)} The image highlights the right side of the upper floor, focusing on the decorative elements on the pillars and the arches. The right most pillar has a thinner middle and wider bottom, and the top pillar decoration has swirly branches and a spread out decoration like a opened flower. The image also shows the inner pillars and the black guard rails.

\paragraph{\texttt{basic} negative:} A grand building with four pillars, three arches, and a detailed facade showcasing mythical creatures and a clock. The building's upper floor boasts five light bulbs, two marble statues, and several arches, while the lower floor features three black guard rails and a streetlight that illuminates the vibrant flowers and trees outside.
\paragraph{\texttt{layout} negative:} A grand building with intricate details, including four pillars, three arches, and a decorative facade now covered in vines and moss, giving it a weathered and natural appearance. The upper floor features five light bulbs, two marble statues, and multiple arches, while the lower floor has three black guard rails and a streetlight.
\paragraph{\texttt{swaps} negative:} A facade with five arches, three pillars, and four light bulbs. The building has a decorative upper floor with two marble statues, multiple black guard rails, and a streetlight. The lower floor has three arches and various decorative elements such as swirling branches, leaves, and faces.

\subsubsection{Submask}

\paragraph{Description:} \textit{Front black street light} - The black street light in front of the ornate building. It has a thin support beam holding it up with little lines going down it. There's curved branches on the beams supporting the lights. Each light has the same design, with little pointy spires or points on its top tip. There's about 6 visible lights on this street light. ...(submask text omitted)...

\paragraph{Selected Summaries (3/8):} \textit{(1)} A black street light stands in front of an ornate building, supported by a thin beam with curved branches. Six lights with pointy spires are visible, with the closest one featuring clear glass panes and a lantern-like shape.\\
\textit{(2)} The front black street light is decorated with curved branches and has a lantern-like shape. The light closest to the image has a clear glass pane with ridges on the edges and a small decorative circle above it.\\
\textit{(3)} This image highlights the unique design of a black street light, with a lantern-like shape and pointed spires on top. The light has a clear glass panel with ridges and a decorative circle, and is supported by a thin beam and curved branches. The image also shows the ornate building in the background.

\paragraph{\texttt{basic} negative:} A white street light stands in front of a modern skyscraper, supported by a thick concrete pillar. Five lights with sleek, angular designs are visible, with the closest one featuring frosted glass panes and a cylindrical shape.
\paragraph{\texttt{layout} negative:} A black street light stands in front of a modern skyscraper, supported by a thick concrete pillar. Six LED lights with sleek, angular designs are visible, with the closest one featuring a frosted glass cover and a cylindrical shape.
\paragraph{\texttt{swaps} negative:} A lantern-like shape stands in front of an ornate building, supported by a clear glass panes with curved branches. Six pointy spires are visible, with the closest one featuring a thin beam and black street light.

%----------------------------------------
\section{Crowdsourcing Methodology}
\label{sec:annotating}

Extending from the overall description for the main task in Section \ref{sec:collection}, we here provide an overview of the precursor qualification tasks, as well as some additional details of the preparation and packaging process for the dataset. The complete set of instructions, as well as code to be able to directly reproduce the collection for this dataset, is available on the project GitHub\footnote{\url{https://github.com/facebookresearch/DCI/tree/main/reproduction/crowdsourcing}}.

Note, we do not include the code for our initial version of the collection task, wherein workers were asked to both select the regions of the image that were worth annotating \textit{and} annotate them in the same pass, as these proved both incredibly time-consuming and often lower quality than using model-based mask generation and allowing workers to filter low-quality masks.

\subsection{Quality Assurance}
High-quality datasets rely on getting solidly-performing crowdworkers, which nowadays can prove to be an adversarial task initially. To remedy this situation, we set up a multi-stage process wherein workers could complete precursor tasks (for which they were compensated) that we could use to determine eligibility in the main task pool.

Stage one of the task asked a few questions about a preset image, wherein we asked workers to provide a few sentences describing the image, then note a few things in the image they might include descriptions of if they were to need to write 1000 meaningful words about the image. After filtering out answers from bots, we manually reviewed answers for quality on the provided description, as well as having included any of a few details in the image that we felt were hard to notice on first pass in either the description or list of additional things they might describe.

Workers with solid fluency in English and solid attention to detail were moved into stage two, wherein they had access to the full task and were eligible to complete it three times. In this stage we evaluated responses for a general understanding of the more complex task interface, and allowed workers who completed this stage or only made minor mistakes to the full tasks. Workers who did make minor mistakes were given feedback to help understanding.

In the last stage, work was audited regularly from each worker to provide feedback about description quality, proper use of disqualifying bad masks, and overall tradeoff between time spent and work completed. We used this information to limit over-contribution from single individual, filter out a few workers that provided decreasing quality over time, and bonus workers who were slower than our target pace but providing exceptional quality.

\subsection{Instruction Details}
Before entering the task, the worker was walked through an example task to familiarize them with the interface and the goal of the task. We additionally provided a few specific scenarios to anticipate common questions, like what to do when two masks are the same, or how to deal with writing for a mask that contains a mask that was already done, or what to do on images that were much simpler than the norm. A complete list of these instructions is available with our released code.

\subsection{Worker Metrics}
Over the course of a month, we made the first qualification round eligible to a large cohort of workers on Mechanical Turk, and fielded over 800 submissions. Of these, roughly 250 workers made it through to the second task.

In the second stage, we hand-reviewed around 600 tasks (up to 3 from each worker from the first stage), and ultimately moved a group of 120 workers to the full task. Of these, around 80 were regular contributors over the course of our collection.

During the main task, we enacted controls to ensure that no worker provided more than 10\% of the currently collected data at any time. We used automated metrics around words per image, words per minute, and unique words per image to act as an overview of worker quality, and hand-reviewed examples that were far outside of our expected bounds for these metrics.

%---------------------------------------
\section{Extended Ablations}
\label{sec:more-ablation}

In order to evaluate ideal use techniques for the Densely Captioned Images dataset for the purpose of fine tuning, we run some ablation experiments and compare to performance on fine tuning on COCO and Localized Narratives. 

\subsection{DCI Fine-tuning ablation methods}

\subsubsection{PickN Caption Training}
In order to make use of all of the available captions for each image, we pick some number of captions from those available. For $N=1$, this just results in selecting one of the captions for each image randomly in each epoch. For $N>1$, we instead provide multiple captions during \textit{both} CLIP loss and negatives loss calculations. In these circumstances, loss is calculated between the \textit{worst} positive and the \textit{best} negative. As each image is supposed to have unique and high-quality captions across the whole set, the expectation here is that any caption for one image should score better than any caption for another.

\subsubsection{Image-based Batching}
When training with $sDCI$ submasks, we have the opportunity to provide the model with exceptionally difficult negatives during the CLIP loss calculation, namely other submasks from the same image. We call this ablation $ImgGroup$ and run it for each $sDCI$ model to determine the impact that hard negatives like these have on CLIP training.

\subsubsection{Negatives Loss}
In order to evaluate the impact that over-weighing negatives during train time has on model performance, we launch one job with 9 times the negatives loss used in our standard experiments. This attempts to evaluate how easily the negative construction techniques in these standard benchmarks could be `gamed'.

\subsubsection{Negatives Selection}
While we have LLM-generated negatives readily available for the $sDCI$ dataset (as described in Section \ref{sec:summarization}, we also wanted to compare to negatives for the $LN$ and $COCO$ datasets. For this we used a \texttt{spacy}-based swapping technique similar to NegCLIP~\citep{aroNegClip} or DAC~\citep{doveh2023dense}, wherein noun phrases, verbs, and adjectives were randomly swapped inside of a given caption to create a negative. We used these \texttt{spacy}-swaps for all $COCO$ and $LN$ runs, but we also include \texttt{spacy}-swap ablations for $DCI$-trained models to compare with the LLM-generated negatives.

\subsection{Datasets and Training methods}

Overall we use the same training parameters outlined in Section \ref{sec:dci-fine-tuning}. We use the following five datasets in our ablations:
\begin{enumerate}
    \item $sDCI_{sub}$: All of the complete images and subimages with LLM generated captions (referred to above as $DCI_{sub<77}$).
    \item $LN$: All images and captions from the COCO subset of Localized Narrations with a CLIP token count under 77 (referred to above as $LN_{COCO<77}$).
    \item $LN_{7805}$: The first 7805 images of $LN$, used to be a same-size comparison to $DCI$-trained models.
    \item $COCO$: All images and captions from the COCO 2017 set.
    \item $COCO_{7805}$ :The first 7805 images of $COCO$, used to be a same-size comparison to $DCI$-trained models.
\end{enumerate}

For each sweep we select the best model as determined by highest score on validation metrics from the same dataset used for training (for instance using $COCO$ valid for a model trained on $COCO$ train).

For $sDCI$, we split the test set into a training set of 7599, a valid set of 98, and a test set of 108. All metrics below are reported on the 108 image test set.

\subsection{Results}

\subsubsection{Aggregate DCI Ablations}

\newcommand{\ApplyGradient}[4]{%
    \ifdim #4 pt > #2 pt
        \pgfmathparse{max(min(100.0*(#4 - #2)/(#3-#2),100.0),0.00)} %
        \xdef\PercentColor{\pgfmathresult}
        \hspace{-0.33em}\cellcolor{green!\PercentColor!yellow} #4\%
    \else
        \pgfmathparse{max(min(100.0*(#2 - #4)/(#2-#1),100.0),0.00)} %
        \xdef\PercentColor{\pgfmathresult}
        \hspace{-0.33em}\cellcolor{red!\PercentColor!yellow} #4\%
    \fi
}

\newcommand{\grada}[1]{\ApplyGradient{46.89}{59.98}{78.61}{#1}}
\newcommand{\gradb}[1]{\ApplyGradient{55.67}{63.18}{68.92}{#1}}
\newcommand{\gradc}[1]{\ApplyGradient{29.60}{47.90}{88.58}{#1}}
\newcommand{\gradd}[1]{\ApplyGradient{37.06}{60.20}{91.40}{#1}}
\newcommand{\grade}[1]{\ApplyGradient{64.80}{81.17}{88.37}{#1}}
\newcommand{\gradf}[1]{\ApplyGradient{63.05}{67.67}{70.42}{#1}}
\newcommand{\gradg}[1]{\ApplyGradient{58.98}{61.95}{71.82}{#1}}
\newcommand{\gradh}[1]{\ApplyGradient{0}{37.82}{64.02}{#1}}
\newcommand{\gradi}[1]{\ApplyGradient{0}{60.12}{97.20}{#1}}
\newcommand{\gradj}[1]{\ApplyGradient{7.46}{10.94}{31.60}{#1}}
\newcommand{\gradk}[1]{\ApplyGradient{0}{23.19}{94.94}{#1}}
\newcommand{\gradl}[1]{\ApplyGradient{0}{67.86}{98.21}{#1}}
\newcommand{\gradm}[1]{\ApplyGradient{0}{39.95}{88.65}{#1}}

\begin{table*}[]
\centering
\footnotesize
\setlength{\tabcolsep}{0\textwidth}
\begin{tabular}{lcc|rrrr|rrr||rr|rr|r|r}
\multicolumn{3}{c|}{Conditions} & \multicolumn{4}{c|}{ARO} & \multicolumn{3}{c||}{VL-Checklist} & \multicolumn{2}{c|}{sDCI All} & \multicolumn{2}{c|}{ All Pick5} & \multicolumn{1}{c}{Base} & \multicolumn{1}{c}{All}\\
Capts & \multicolumn{1}{l}{NL} & \multicolumn{1}{l|}{Batch} & \multicolumn{1}{c}{VG-R} & \multicolumn{1}{c}{VG-A} & \multicolumn{1}{c}{COCO} & \multicolumn{1}{c|}{FLICKR} & \multicolumn{1}{c}{Object} & \multicolumn{1}{c}{Attribute} & \multicolumn{1}{c||}{Relation} & \multicolumn{1}{c}{SCM} & \multicolumn{1}{c|}{Neg} & \multicolumn{1}{c}{SCM} & \multicolumn{1}{c|}{Neg} & \multicolumn{1}{c|}{Neg} & \multicolumn{1}{c}{H-Neg}
\\ \hline
First & 9 & Rand & \grada{75.34} & \gradb{62.72} & \gradc{84.33} & \gradd{87.86} & \grade{76.99} & \gradf{67.77} & \gradg{68.27} & \gradh{51.71} & \textbf{\gradi{97.20}} & \gradj{8.69} & \gradk{79.41} & \textbf{\gradl{98.21}} & \gradm{88.30} \\
First & 9 & Group & \grada{74.82} & \gradb{58.26} & \gradc{85.05} & \gradd{88.70} & \grade{69.44} & \gradf{68.65} & \gradg{68.93} & \gradh{54.38} & \gradi{96.79} & \gradj{9.44} & \gradk{79.69} & \gradl{94.64} & \textbf{\gradm{88.65}}\\
Pick1 & 9 & Rand & \grada{76.04} & \gradb{65.51} & \gradc{86.69} & \gradd{89.72} & \grade{73.35} & \gradf{65.04} & \gradg{69.48} & \gradh{44.12} & \gradi{95.55} & \gradj{12.59} & \gradk{93.57} & \gradl{93.75} & \gradm{82.97}\\ 
Pick1 & 9 & Group & \grada{76.41} & \gradb{63.71} & \gradc{82.89} & \gradd{88.04} & \grade{72.68} & \gradf{65.44} & \gradg{63.33} & \gradh{50.14} & \gradi{95.62} & \gradj{19.36} & \gradk{94.05} & \gradl{92.86} & \gradm{84.06}\\ 
Pick5 & 9 & Rand & \textbf{\grada{78.61}} & \gradb{65.65} & \gradc{86.83} & \textbf{\gradd{91.40}} & \grade{68.61} & \gradf{66.10} & \textbf{\gradg{71.82}} & \gradh{39.33} & \gradi{96.10} & \gradj{7.46} & \textbf{\gradk{94.94}} & \textbf{\gradl{98.21}} & \gradm{85.64}\\
Pick5 & 9 & Group & \grada{75.65} & \textbf{\gradb{68.92}} & \gradc{85.45} & \gradd{90.26} & \grade{70.17} & \gradf{65.09} & \gradg{63.80} & \gradh{43.23} & \gradi{96.17} & \gradj{11.56} & \gradk{94.32} & \gradl{96.43} & \gradm{83.79}\\
\hline
First & 1 & Rand & \grada{64.00} & \gradb{60.79} & \gradc{83.93} & \gradd{87.46} & \grade{76.73} & \gradf{68.75} & \gradg{63.53} & \gradh{54.17} & \gradi{96.17} & \gradj{14.64} & \gradk{75.92} & \gradl{96.43} & \gradm{85.91}\\
First & 1 & Group & \grada{73.09} & \gradb{62.60} & \gradc{79.54} & \gradd{84.84} & \grade{72.64} & \gradf{69.37} & \gradg{61.68} & \gradh{61.76} & \gradi{95.76} & \gradj{19.63} & \gradk{76.81} & \gradl{97.32} & \gradm{85.43}\\
Pick1 & 1 & Rand & \grada{76.23} & \gradb{67.56} & \textbf{\gradc{88.58}} & \gradd{91.30} & \grade{80.71} & \gradf{68.69} & \gradg{70.12} & \gradh{50.48} & \gradi{94.39} & \gradj{19.02} & \gradk{88.85} & \gradl{95.54} & \gradm{81.40}\\ 
Pick1 & 1 & Group & \grada{64.97} & \gradb{65.05} & \gradc{80.32} & \gradd{87.62} & \grade{75.01} & \gradf{66.70} & \gradg{60.75} & \gradh{58.14} & \gradi{94.60} & \gradj{26.81} & \gradk{89.88} & \gradl{96.43} & \gradm{83.58}\\ 
Pick5 & 1 & Rand & \grada{72.56} & \gradb{61.32} & \gradc{78.00} & \gradd{83.68} & \grade{77.64} & \gradf{66.74} & \gradg{66.65} & \gradh{49.93} & \gradi{94.19} & \gradj{18.60} & \gradk{86.87} & \gradl{95.54} & \gradm{80.78}\\
Pick5 & 1 & Group & \grada{57.96} & \gradb{61.49} & \gradc{78.27} & \gradd{84.06} & \grade{64.80} & \gradf{63.05} & \gradg{61.73} & \gradh{51.44} & \gradi{94.66} & \gradj{19.97} & \gradk{90.63} & \gradl{93.75} & \gradm{81.81}\\
\hline
First & 0 & Rand & \grada{55.61} & \gradb{55.83} & \gradc{29.60} & \gradd{39.22} & \grade{80.70} & \gradf{67.71} & \gradg{62.10} & \gradh{57.25} & \gradi{74.21} & \gradj{22.23} & \gradk{30.30} & \gradl{80.36} & \gradm{67.58} \\
First & 0 & Group & \grada{49.91} & \gradb{59.04} & \gradc{41.68} & \gradd{50.90} & \grade{73.11} & \gradf{67.35} & \gradg{60.03} & \gradh{63.13} & \gradi{75.44} & \gradj{27.84} & \gradk{33.11} & \gradl{81.25} & \gradm{69.49}\\
Pick1 & 0 & Rand & \grada{57.34} & \gradb{61.98} & \gradc{39.36} & \gradd{44.62} & \textbf{\grade{88.37}} & \textbf{\gradf{70.42}} & \gradg{61.28} & \gradh{56.77} & \gradi{74.08} & \gradj{23.73} & \gradk{34.54} & \gradl{83.93} & \gradm{64.43} \\ 
Pick1 & 0 & Group & \grada{50.88} & \gradb{56.67} & \gradc{46.62} & \gradd{51.82} & \grade{76.83} & \gradf{67.41} & \gradg{58.98} & \textbf{\gradh{64.02}} & \gradi{71.55} & \textbf{\gradj{31.60}} & \gradk{35.16} & \gradl{78.57} & \gradm{66.28}\\ 
Pick5 & 0 & Rand & \grada{53.64} & \gradb{62.39} & \gradc{36.70} & \gradd{43.78} & \grade{81.33} & \gradf{69.79} & \gradg{61.40} & \gradh{56.77} & \gradi{75.99} & \gradj{22.16} & \gradk{30.78} & \gradl{82.14} & \gradm{68.60} \\
Pick5 & 0 & Group & \grada{46.89} & \gradb{55.67} & \gradc{32.05} & \gradd{37.06} & \grade{75.24} & \gradf{68.98} & \gradg{63.35} & \gradh{61.97} & \gradi{77.84} & \gradj{26.88} & \gradk{32.35} & \gradl{79.46} & \gradm{73.05} \\
\hline \hline
\rowcolor{yellow} \multicolumn{3}{c|}{CLIP Baseline}  &  \grada{59.98} & \gradb{63.18} & \gradc{47.9} & \gradd{60.2} & \grade{81.17} & \gradf{67.67} & \gradg{61.95} & \gradh{37.82} & \gradi{60.12} & \gradj{10.94} & \gradk{23.19} & \gradl{67.86} & \gradm{39.95}\\
\end{tabular}
\normalsize
\caption{Full $sDCI$ ablation analysis against all benchmarks. Cells are colored in comparison to the CLIP Baseline.}
\label{tab:ablate}
\end{table*}

A complete table of our sDCI ablations can be seen in Table \ref{tab:ablate}, however it is more dense information than is likely useful in seeing overall trends. Instead, we aggregate over the different ablation methods.

\begin{table*}[]
\centering
\begin{tabular}{l|rrrr|rrr}
&  \multicolumn{4}{c|}{ARO} & \multicolumn{3}{c}{VL-Checklist} \\
\multicolumn{1}{c|}{Ablation} & \multicolumn{1}{c}{VG-R} & \multicolumn{1}{c}{VG-A} & \multicolumn{1}{c}{COCO} & \multicolumn{1}{c|}{FLICKR} & \multicolumn{1}{c}{Object} & \multicolumn{1}{c}{Attribute} & \multicolumn{1}{c}{Relation} \\ \hline
Rand & \textbf{67.71}\% & \textbf{62.64}\% & \textbf{68.22}\% & 73.23\% & \textbf{77.71}\% & \textbf{67.89}\% & \textbf{66.07}\% \\
ImgGroup & 63.40\% & 61.27\% & 67.99\% & \textbf{73.70}\% & 72.21\% & 66.89\% & 62.51\% \\
\hline
Neg Loss 9 & \textbf{76.15}\% & \textbf{64.13}\% & \textbf{85.21}\% & \textbf{89.33}\% & 71.87\% & 66.35\% & \textbf{67.61}\% \\
Neg Loss 1 & 68.14\% & 63.14\% & 81.44\% & 86.49\% & 74.59\% & 67.22\% & 64.08\% \\
Neg Loss 0 & 52.38\% & 58.60\% & 37.67\% & 44.57\% & \textbf{78.43}\% & \textbf{68.61}\% & 61.19\% \\
\hline
First & 65.46\% & 59.87\% & 67.36\% & 73.16\% & 74.94\% & \textbf{68.27}\% & 64.09\% \\
Pick1 & \textbf{66.98}\% & \textbf{63.41}\% & \textbf{70.74}\% & \textbf{75.52}\% & \textbf{76.99}\% & 67.28\% & 63.99\% \\
Pick5 & 64.22\% & 62.57\% & 66.22\% & 71.71\% & 72.97\% & 66.63\% & \textbf{64.79}\% \\
\end{tabular}
\caption{$sDCI$ Grouped ablation against standard benchmarks. Results from Table \ref{tab:ablate} are averaged across different ablations.}
\label{tab:ablate-aro-vlc}
\end{table*}

In Table \ref{tab:ablate-aro-vlc}, for ARO and VL-C we observe that generally constructing batches from the entire dataset rather than from masks in the same image is better for performance. 

We also see that using \textit{Pick1} is the most effective method for making use out of the LLM-summarizations of captions we've created, as it allows the model to see more of the related text (better than \textit{first}) without potentially penalizing it for situations where two related overlapping captions are provided at the same time (which may be an issue with \textit{Pick5}).

We do observe that using high negatives loss is very effective at gaining additional performance on most of these metrics, however this does not include VL-Object and VL-Attribute, likely due to our method of constructing negatives not correlating very well with the types of negatives created in these tests. While this provides great scores, it mostly just provides evidence towards the aforementioned issues with these types of evaluations \cite{lin2023visualgptscore}.

\begin{table*}[]
\centering
\begin{tabular}{l|rr|rr|r|r}
& \multicolumn{2}{c|}{All}                               & \multicolumn{2}{c|}{All Pick5}                         & \multicolumn{1}{c|}{Base} & \multicolumn{1}{c}{All}       \\
Ablation & \multicolumn{1}{c}{SCM}  & \multicolumn{1}{c|}{Neg}   & \multicolumn{1}{c}{SCM}  & \multicolumn{1}{c|}{Neg}   & \multicolumn{1}{c|}{Neg}  & \multicolumn{1}{c}{Hard Negs} \\ \hline
Rand & 51.17\% & 88.65\% & 16.57\% & 68.35\% & \textbf{91.57}\% & 78.40\% \\
ImgGroup & \textbf{56.47}\% & \textbf{88.71}\% & \textbf{21.45}\% & \textbf{69.56}\% & 90.08\% & \textbf{79.57}\% \\
\hline
NL 9 & 47.15\% & 96.24\% & 11.52\% & \textbf{89.33}\% & 95.68\% & \textbf{85.57}\% \\
NL 1 & 54.32\% & \textbf{94.96}\% & 19.78\% & 84.83\% & \textbf{95.84}\% & 83.15\% \\
NL 0 & \textbf{59.99}\% & 74.85\% & \textbf{25.74}\% & 32.71\% & 80.95\% & 68.24\% \\
\hline
First & \textbf{57.07}\% & \textbf{89.26}\% & 17.08\% & 62.54\% & \textbf{91.37}\% & \textbf{80.89}\% \\
Pick1 & 53.95\% & 87.63\% & \textbf{22.19}\% & \textbf{72.68}\% & 90.18\% & 77.12\% \\
Pick5 & 50.45\% & 89.16\% & 17.77\% & 71.65\% & 90.92\% & 78.95\% \\

\end{tabular}
\caption{$sDCI$ Grouped ablation against the 112 heldout sDCI test images. Results from Table \ref{tab:ablate} are averaged across different ablations.}
\label{tab:ablate-dci}
\end{table*}

In Table \ref{tab:ablate-dci}, we observe a slightly different story. Here $ImgGroup$ appears to be the best method for selecting images, only performing worse on the Base Neg test. This is expected though, as the test set is constructed with sequential examples in a manner similar to the $ImgGroup$ ablation, which doesn't have any effect on Base Neg given there are no submasks to deal with in that test.

No individual setting for negatives loss performs best on all metrics, however it is unsurprising that negatives loss 0 results in the highest performance on Subcrop-Caption Matching tasks (given their similarity to the CLIP-style learning objective), and not as well as negatives-trained models on negatives tasks.

The PickN ablations are somewhat surprising, as \textit{first} captions generally performed the best overall, and Pick1 outperformed training on Pick5 when testing on Pick5.

\subsubsection{DCI fine-tuning performance}

\begin{table*}[]
\centering
\begin{tabular}{llll|rrrrrr}
\multicolumn{4}{c|}{Training Parameters}  & \multicolumn{2}{c}{All}                               & \multicolumn{2}{c}{All Pick5}                         & \multicolumn{1}{c}{Base} & \multicolumn{1}{c}{All}       \\
Dataset              & Captions & Negatives & Batching & \multicolumn{1}{c}{SCM}  & \multicolumn{1}{c}{Neg}   & \multicolumn{1}{c}{SCM}  & \multicolumn{1}{c}{Neg}   & \multicolumn{1}{c}{Neg}  & \multicolumn{1}{c}{Hard Negs} \\ \hline
$sDCI_{7805}$ & First & LLM & Rand & 38.30\% & 84.61\% & 9.10\% & 69.22\% & 92.86\% & 76.54\% \\
$sDCI$ & Pick1 & LLM & ImgGroup & 58.14\% & \textbf{94.60\%} & 26.81\% & \textbf{89.88\%} & 96.43\% & 83.58\% \\
$sDCI$ & Pick1 & LLM & Rand & 50.48\% & 94.39\% & 19.02\% & 88.85\% & 95.54\% & 81.40\% \\
$sDCI$ & First & LLM & ImgGroup & 61.76\% & 95.76\% & 19.63\% & 76.81\% & \textbf{97.32\%} & 85.43\% \\
$sDCI$ & First & LLM & Rand & 54.17\% & 96.17\% & 14.64\% & 75.92\% & 96.43\% & \textbf{85.91\%} \\
$sDCI$ & First & Spacy & Rand & 55.13\% & 87.35\% & 19.08\% & 59.30\% & 89.29\% & 75.85\% \\
\hline
$sDCI$ & Pick1 & None  & ImgGroup & \textbf{64.02}\% & 71.55\% & \textbf{31.60\%} & 35.15\% & 78.57\% & 66.28\% \\
$sDCI$ & Pick1 & None & Rand & 56.77\% & 74.08\% & 23.73\% & 34.54\% & 83.93\% & 64.43\% \\
$sDCI$ & First & None & ImgGroup & 63.13\% & 75.44\% & 27.84\% & 33.11\% & 81.25\% & 69.49\%\\
$sDCI$ & First & None & Rand & 57.25\% & 74.21\% & 22.23\% & 30.30\% & 80.36\% & 67.58\% \\
\hline \hline
$LN$ & First & Spacy & Rand & 37.82\% & 76.95\% & 9.37\% & 46.31\% & 86.61\% & 63.20\% \\
$LN_{7805}$ & First & Spacy & Rand & 34.27\% & 75.58\% & 7.73\% & 37.82\% & 83.04\% & 61.63\% \\
$LN$ & First & None & Rand & 41.45\% & 58.82\% & 12.72\% & 21.75\% & 80.36\% & 53.42\% \\
$COCO$ & First & Spacy & Rand & 40.97\% & 79.21\% & 12.65\% & 52.74\% & 91.07\% & 64.71\% \\ 
$COCO_{7805}$ & First & Spacy & Rand & 38.51\% & 79.75\% & 11.70\% & 55.27\% & 86.61\% & 64.16\% \\ 
$COCO$ & First & None & Rand & 42.00\% & 61.35\% & 13.95\% & 21.41\% & 82.14\% & 52.60\% \\
\hline \hline
\multicolumn{4}{c|}{CLIP Baseline }& 37.82\% &	60.12\% &	10.94\% &	23.19\% &	67.86\% &	39.95\% \\
\multicolumn{4}{c|}{$DAC_{LLM}$} & 36.87\% & 81.12\% &  8.00\% & 35.91\% & 86.61\% & 70.66\% \\
\multicolumn{4}{c|}{$DAC_{SAM}$} &  36.46\% & 84.40\% & 6.91\% & 40.83\% & 89.29\% & 73.94\%\\
\end{tabular}
\caption{Dense Captions test results. We compare $DCI$-trained models to models trained on Localized Narratives and COCO datasets, as well as to baselines.}
\label{tab:dci-test-LN-COCO}
\end{table*}

In Table \ref{tab:dci-test-LN-COCO} we report our different ablations performance compared to the LN and COCO baselines, as well as CLIP and DAC as comparison points. We expect sDCI models to outperform all other baselines, given the test set is out-of-distribution for the other models. Still, there are some interesting observations available in this table.

First, for sDCI, Localized Narrations, and COCO, only training on the smallest subset of images (7800) and without using masks, DCI ends up performing only as well as COCO for Subcrop-Caption matching, but both outperform Localized Narratives by a noticeable margin. This may point to data in Localized Narratives generally being less sample-efficient than baseline COCO captions. 

Second, moving to LLM-based captions increases performance on all negatives at the expense of performance on Subcrop-Caption masking. This would imply that the captions generated by LLMs may actually be overfitting to the test task to the detriment of performance on other metrics. 

Third, training directly on negatives as a method of improving models' vision-language understanding universally decreases performance on Subcrop-Caption masking, a task that also definitely takes strong vision-language understanding. This is seen regardless of the training dataset used, or of any other sDCI ablation involved.

\subsubsection{Linear Transferability}
We evaluate a subset of all of our models, trained on each of our candidate datasets, on the Elevater \cite{li2022elevater} benchmark to determine linear transferability for our models. We expect some degradation given the relatively small size of our training datasets. We also evaluate them on zero-shot ImageNet specifically.

Overall, in table \ref{tab:dci-elevater} we observe a slight degradation in linear probe performance on their included datasets across all shots. In our ImageNet zero-shot evaluation reported in Table \ref{tab:in-zero-shot}, we note that sDCI trained without negatives suffers much less degradation compared to with negatives, ending up with comparable performance to the DAC models.

\begin{table}[]
\centering
\setlength{\tabcolsep}{2pt}
\begin{tabular}{l|c}
\multicolumn{2}{c}{ImageNet 0-Shot} \\ 
Model & Valid. Accuracy\\
\hline
CLIP (baseline)  & 60.96\% \\
$sDCI_{P1}$ & 42.51\% \\
$sDCI_{P1NL0}$ & 51.44\%\\
DAC-LLM & 52.65\%\\
DAC-SAM & 53.43\%\\
\hline
\end{tabular}
\caption{ImageNet zero-shot.}
\label{tab:in-zero-shot}
\end{table}

\begin{table}[]
\centering
\setlength{\tabcolsep}{2pt}
\begin{tabular}{l|ccccc}
 &  \multicolumn{5}{c}{Elevater N-Shot} \\ 
Model &  \multicolumn{1}{c}{0} & \multicolumn{1}{c}{5} &  \multicolumn{1}{c}{20} & \multicolumn{1}{c}{50} & \multicolumn{1}{c}{Full} \\ \hline
$sDCI_{P1}$  & 45.10\% &60.78\% &69.54\% &72.91\% &77.49\% \\
$sDCI_{P1NL0}$ & 51.29\% &61.59\% &70.80\% &73.39\% &77.92\%\\
\hline
CLIP & 55.59\% &64.85\% &71.90\% &74.38\% &78.96\% \\

\end{tabular}
\caption{Elevater scores for linear probe across 20 benchmark datasets}
\label{tab:dci-elevater}
\end{table}

%----------------------------------------
\section{Additional Selected Examples}
\label{sec:more-examples}

We include a few additional examples from the DCI dataset, selected from a random subset of 20 instances to highlight certain elements of the dataset. In each we share a subset of the masks available per image.

\begin{figure*}
  \centering
  \includegraphics[width=0.95\textwidth]{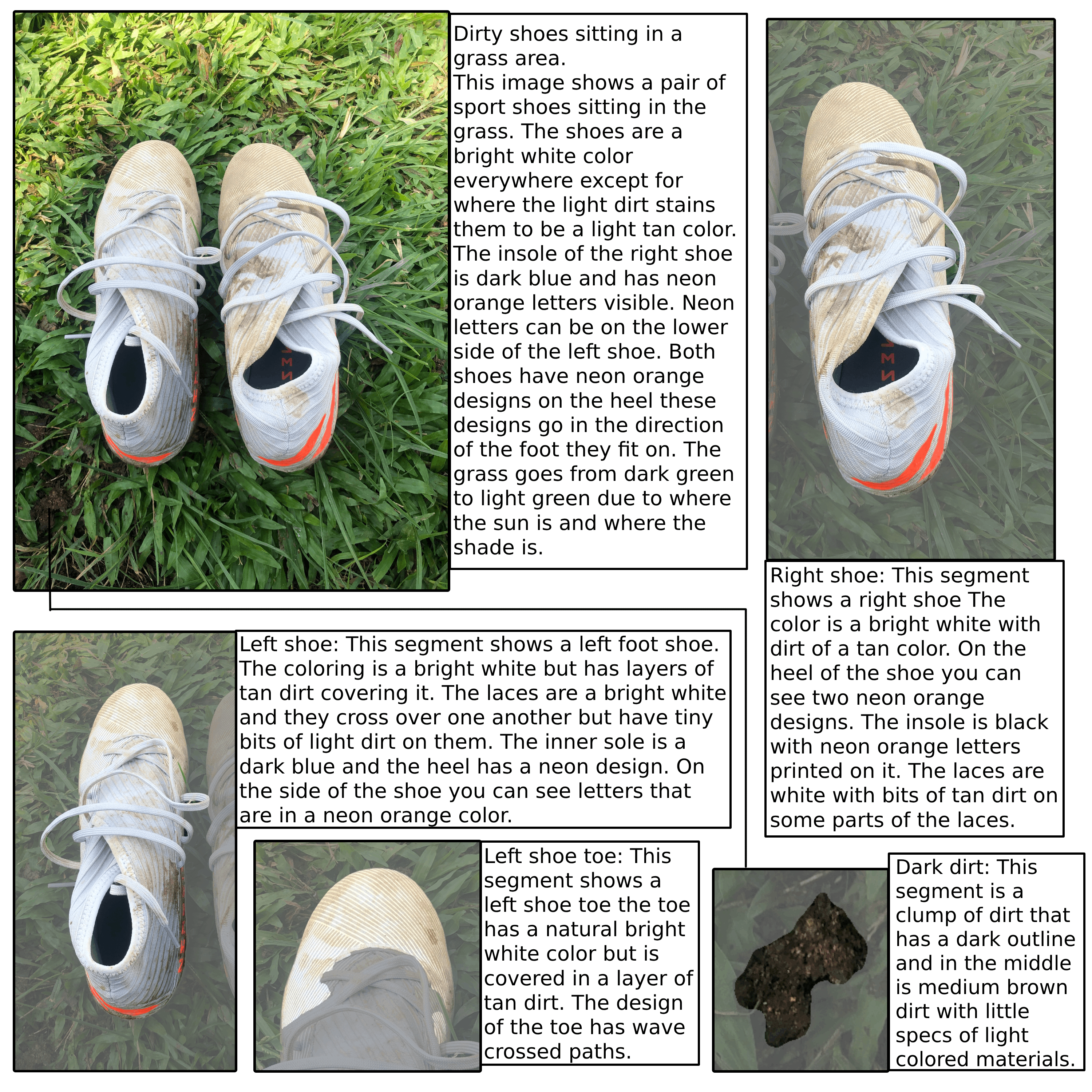}
  \caption{One example from the Densely Captioned Images dataset, highlighting how in-depth descriptions are provided even for relatively simple scenes.}
  \label{fig:dc_example2}
\end{figure*}

Figure \ref{fig:dc_example2} shows the level in-depth that the descriptions go to. Despite only being an image of shoes on some grass, DCI contains descriptions down to the details of the crossing pattern on the toecap or the specks of light colored materials in a clump of dirt on the ground.

\begin{figure*}
  \centering
  \includegraphics[width=0.95\textwidth]{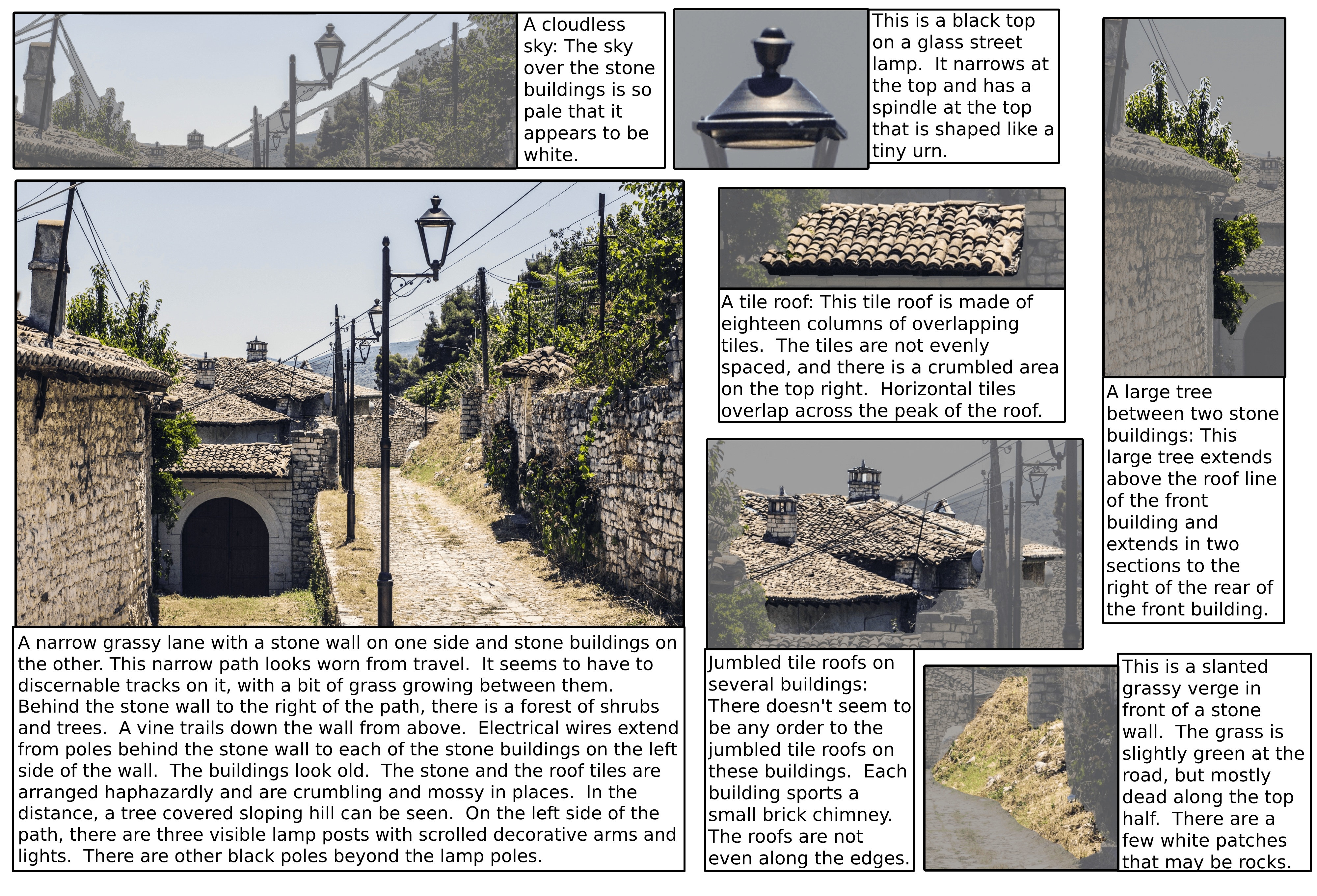}
  \caption{One example from the Densely Captioned Images dataset, highlighting how text is still highly aligned even with complex masks.}
  \label{fig:dc_example3}
\end{figure*}

Figure \ref{fig:dc_example3} contains a fairly complex scene of various tiled stone buildings and light fixtures, however the description is able to identify a tree that is mostly obscured by a foreground building, as well as be in-depth enough to describe the shape at the top of a lamppost in the image as a ``tiny urn''.

\begin{figure*}
 \centering
 \includegraphics[width=0.95\textwidth]{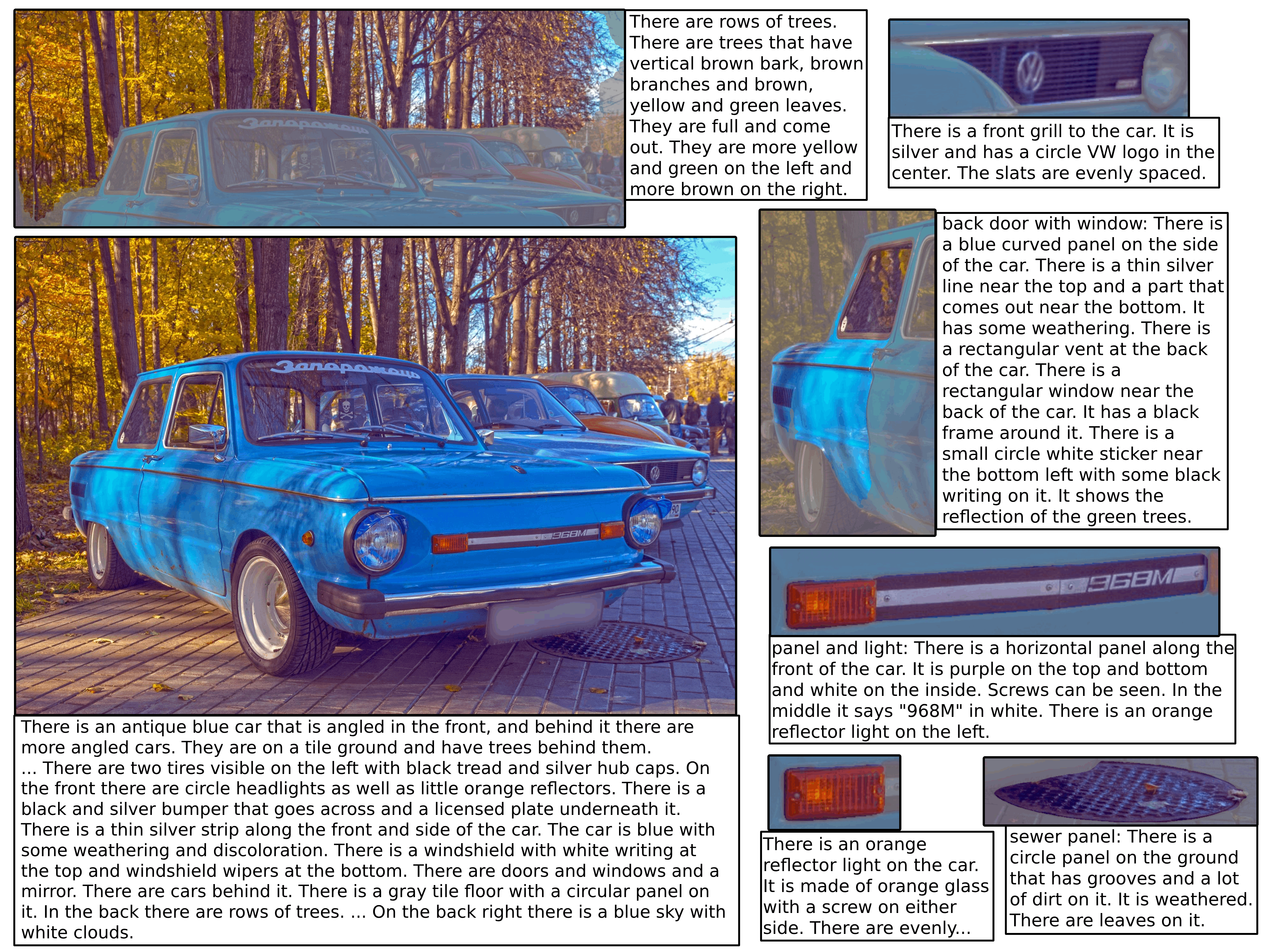}
 \caption{One example from the Densely Captioned Images dataset, highlighting the high resolution of the text in alignment with the image, down to details of the stickers on car windows or screws on the reflectors.}
 \label{fig:dc_example4}
\end{figure*}

Figure \ref{fig:dc_example4} stands as another example of retaining useful and aligned information even when there's a high amount of potential complexity in the image. While "An antique blue car in front of a row of trees" may be a standard caption for this image, we instead have details of the orientation of the car, details of what is visible in frame, and the resolution of the text goes all the way down to the small circular sticker on the rear passenger side door, or the screws on the reflector on the front bumper.

\begin{figure*}
 \centering
 \includegraphics[width=0.95\textwidth]{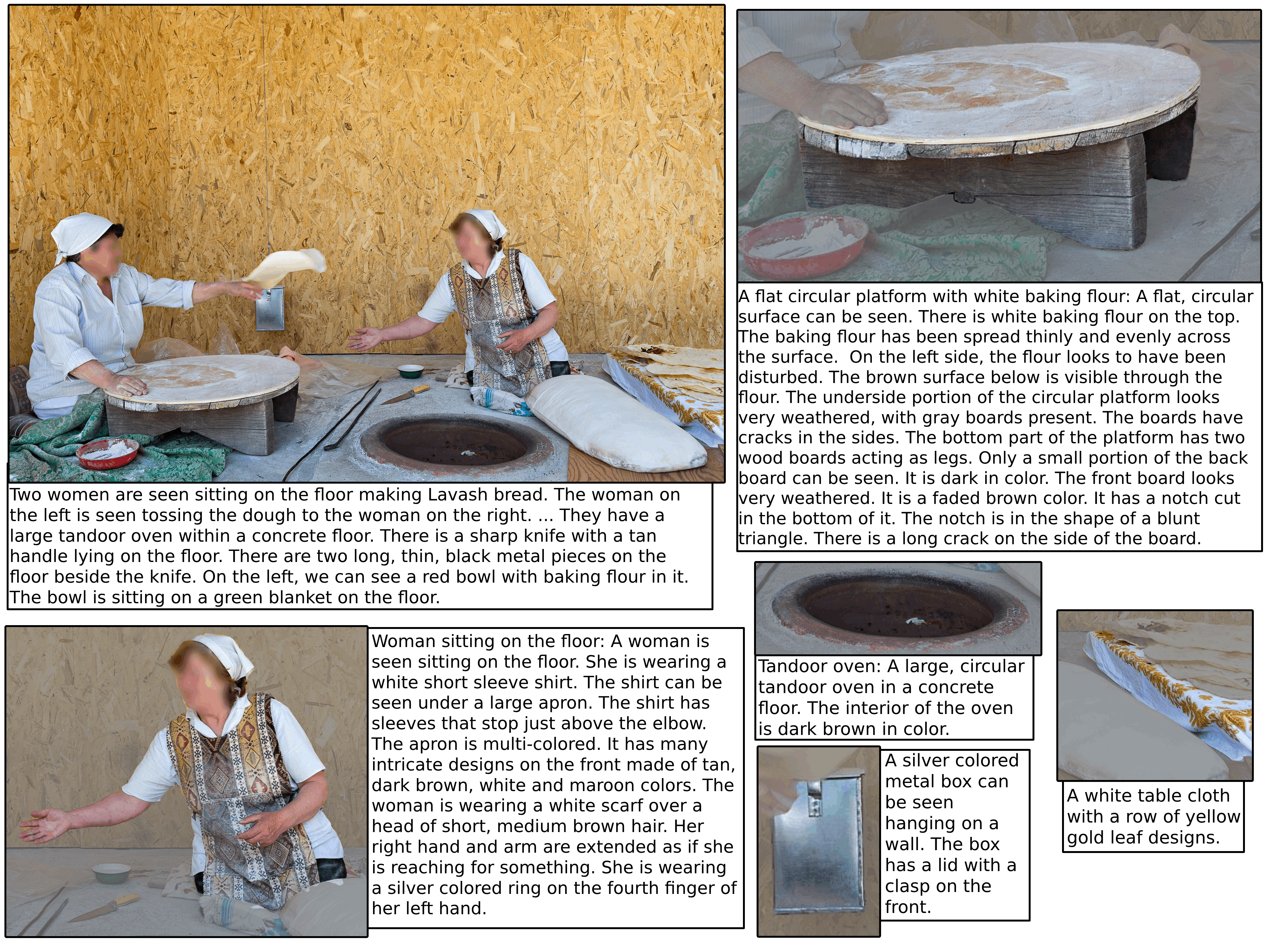}
 \caption{One example from the Densely Captioned Images dataset, displaying a scene with a complex interaction. The aligned description captures the action itself, alongside in-depth details like the clasp on the box on the wall, or the hand of one of the women pictured.}
 \label{fig:dc_example5}
\end{figure*}

Figure \ref{fig:dc_example5} displays a more active scene of two women cooking bread, however it still captures in-depth descriptions of everything contained in the image including flour spread on a table, ornamental details on a tablecloth, and additionally the placement of a bowl and a knife that were not captured in their own masks, but still were successfully annotated.

%----------------------------------------
\section{Evaluating additional baselines on sDCI}

We evaluate a large quantity of available VLMs from RN50, ViT, roberta, convnext, and coca architectures on the DCI test set. Results can be seen in Table \ref{tab:dci-full-test}.

Generally we observe that larger models perform better, but no model excels at all tasks. In contrast to what is observed for models trained with negatives, performance from these pretraining objectives is positively correlated between Subcrop-Caption Matching and detecting negatives.

The most performant model on SCM averaged between the two tasks is coca\_ViT-L-14 mscoco\_finetuned\_laion2b\_s13b\_b90k. The most performant model in average across negatives tasks is ViT-g-14 laion2b\_s34b\_b88k

\begin{table*}[ht]
\centering
\footnotesize
\setlength{\tabcolsep}{3pt}
\begin{tabular}{ll|rrrrrr}
\multicolumn{2}{c|}{Training Parameters}  & \multicolumn{2}{c}{All}                               & \multicolumn{2}{c}{All Pick5}                         & \multicolumn{1}{c}{Base} & \multicolumn{1}{c}{All}       \\
\multicolumn{1}{c|}{Arch} & \multicolumn{1}{c|}{Dataset} & \multicolumn{1}{c}{SCM}  & \multicolumn{1}{c}{Neg}   & \multicolumn{1}{c}{SCM}  & \multicolumn{1}{c}{Neg}   & \multicolumn{1}{c}{Neg}  & \multicolumn{1}{c}{Hard Negs} \\ \hline
\hline \hline
RN50 & yfcc15m & 39.95\% & 55.85\% & 6.72\% & 19.30\% & 71.63\% & 55.35\% \\
RN50 & cc12m & 41.32\% & 48.79\% & 8.65\% & 19.69\% & 65.38\% & 49.72\% \\
RN50-quickgelu & openai & 41.54\% & 62.20\% & 11.58\% & 23.51\% & 72.24\% & 54.00\% \\
RN50-quickgelu & yfcc15m & 40.08\% & 55.89\% & 6.63\% & 19.06\% & 70.51\% & 55.14\% \\
RN50-quickgelu & cc12m & 41.97\% & 47.70\% & 9.18\% & 17.98\% & 65.41\% & 48.63\% \\
RN101 & yfcc15m & 39.72\% & 55.85\% & 7.02\% & 19.73\% & 72.01\% & 54.88\% \\
RN101-quickgelu & openai & 40.44\% & 62.70\% & 10.53\% & 19.48\% & 76.52\% & 56.22\% \\
RN101-quickgelu & yfcc15m & 39.88\% & 55.51\% & 7.14\% & 19.31\% & 71.07\% & 54.47\% \\
ViT-B-32 & openai & 40.06\% & 60.79\% & 11.23\% & 24.12\% & 67.56\% & 41.34\% \\
ViT-B-32 & laion400m\_e31 & 45.44\% & 62.96\% & 14.75\% & 22.29\% & 79.33\% & 57.27\% \\
ViT-B-32 & laion400m\_e32 & 45.31\% & 62.95\% & 14.65\% & 22.07\% & 79.13\% & 57.27\% \\
ViT-B-32 & laion2b\_e16 & 45.59\% & 60.13\% & 15.34\% & 20.23\% & 78.60\% & 56.26\% \\
ViT-B-32 & laion2b\_s34b\_b79k & 45.56\% & 60.12\% & 15.30\% & 20.58\% & 78.10\% & 56.28\% \\
ViT-B-32 & datacomp\_xl\_s13b\_b90k & 45.62\% & 59.92\% & 15.09\% & 20.24\% & 78.17\% & 56.19\% \\
ViT-B-32 & datacomp\_m\_s128m\_b4k & 40.29\% & 58.53\% & 8.93\% & 19.93\% & 73.43\% & 55.70\% \\
ViT-B-32 & commonpool\_m\_clip\_s128m\_b4k & 41.58\% & 58.35\% & 10.60\% & 19.63\% & 71.95\% & 54.76\% \\
ViT-B-32 & commonpool\_m\_laion\_s128m\_b4k & 40.25\% & 59.00\% & 8.92\% & 20.45\% & 71.89\% & 55.81\% \\
ViT-B-32 & commonpool\_m\_image\_s128m\_b4k & 41.14\% & 58.33\% & 9.95\% & 20.28\% & 73.66\% & 55.54\% \\
ViT-B-32 & commonpool\_m\_basic\_s128m\_b4k & 40.86\% & 58.32\% & 9.81\% & 19.54\% & 73.41\% & 55.24\% \\
ViT-B-32 & datacomp\_s\_s13m\_b4k & 28.98\% & 55.47\% & 1.31\% & 18.44\% & 61.73\% & 54.16\% \\
ViT-B-32 & commonpool\_s\_clip\_s13m\_b4k & 33.96\% & 59.22\% & 4.62\% & 21.46\% & 68.17\% & 56.34\% \\
ViT-B-32 & commonpool\_s\_laion\_s13m\_b4k & 28.99\% & 55.50\% & 1.37\% & 18.45\% & 61.43\% & 54.27\% \\
ViT-B-32 & commonpool\_s\_image\_s13m\_b4k & 28.98\% & 55.47\% & 1.33\% & 18.31\% & 61.73\% & 54.16\% \\
ViT-B-32 & commonpool\_s\_basic\_s13m\_b4k & 32.31\% & 57.03\% & 3.39\% & 19.01\% & 64.84\% & 54.89\% \\
ViT-B-32 & commonpool\_s\_s13m\_b4k & 32.45\% & 58.97\% & 4.80\% & 22.63\% & 67.50\% & 55.64\% \\
ViT-B-32-quickgelu & openai & 40.06\% & 60.79\% & 11.22\% & 24.18\% & 67.56\% & 41.34\% \\
ViT-B-32-quickgelu & laion400m\_e31 & 45.19\% & 61.19\% & 14.75\% & 21.76\% & 78.27\% & 55.67\% \\
ViT-B-32-quickgelu & laion400m\_e32 & 45.08\% & 61.14\% & 14.83\% & 21.88\% & 78.45\% & 55.60\% \\
ViT-B-32-quickgelu & metaclip\_400m & 44.27\% & 60.73\% & 12.81\% & 20.28\% & 77.68\% & 57.23\% \\
ViT-B-32-quickgelu & metaclip\_fullcc & 44.63\% & 59.75\% & 14.05\% & 20.61\% & 79.09\% & 56.34\% \\
ViT-B-16 & openai & 39.33\% & 61.40\% & 11.30\% & 26.02\% & 70.98\% & 51.96\% \\
ViT-B-16 & laion400m\_e31 & 44.66\% & 61.46\% & 14.66\% & 22.60\% & 78.07\% & 56.09\% \\
ViT-B-16 & laion2b\_s34b\_b88k & 45.28\% & 60.81\% & 15.28\% & 20.60\% & 77.72\% & 57.44\% \\
ViT-B-16 & datacomp\_xl\_s13b\_b90k & 45.31\% & 59.96\% & 15.39\% & 20.42\% & 76.76\% & 56.48\% \\
ViT-B-16 & datacomp\_l\_s1b\_b8k & 44.41\% & 59.45\% & 14.12\% & 20.32\% & 76.99\% & 55.91\% \\
ViT-B-16 & commonpool\_l\_clip\_s1b\_b8k & 44.06\% & 60.71\% & 14.00\% & 20.73\% & 76.82\% & 55.88\% \\
ViT-B-16 & commonpool\_l\_laion\_s1b\_b8k & 44.53\% & 60.45\% & 14.22\% & 20.40\% & 76.66\% & 56.24\% \\
ViT-B-16 & commonpool\_l\_image\_s1b\_b8k & 44.02\% & 58.72\% & 13.86\% & 19.56\% & 76.50\% & 55.23\% \\
ViT-B-16 & commonpool\_l\_text\_s1b\_b8k & 44.48\% & 60.21\% & 14.02\% & 20.67\% & 78.46\% & 56.91\% \\
ViT-B-16 & commonpool\_l\_basic\_s1b\_b8k & 44.19\% & 59.99\% & 13.89\% & 19.79\% & 77.98\% & 56.47\% \\
ViT-B-16 & commonpool\_l\_s1b\_b8k & 43.11\% & 58.31\% & 13.38\% & 19.10\% & 75.81\% & 54.20\% \\
ViT-B-16-quickgelu & metaclip\_400m & 43.88\% & 61.31\% & 13.28\% & 21.47\% & 79.17\% & 57.95\% \\
ViT-L-14 & openai & 38.09\% & 59.76\% & 11.05\% & 24.82\% & 69.15\% & 52.42\% \\
ViT-L-14 & laion400m\_e31 & 44.09\% & 63.61\% & 14.89\% & 21.92\% & 80.22\% & 57.90\% \\
ViT-L-14 & laion400m\_e32 & 44.09\% & 63.43\% & 14.86\% & 21.65\% & 80.04\% & 57.78\% \\
ViT-L-14 & laion2b\_s32b\_b82k & 45.19\% & 58.82\% & 15.25\% & 19.82\% & 76.96\% & 56.73\% \\
ViT-L-14 & datacomp\_xl\_s13b\_b90k & 44.37\% & 63.19\% & 15.37\% & 22.43\% & 79.99\% & 59.45\% \\
ViT-L-14 & commonpool\_xl\_clip\_s13b\_b90k & 44.63\% & 61.28\% & 15.52\% & 21.37\% & 77.50\% & 58.05\% \\
ViT-L-14 & commonpool\_xl\_laion\_s13b\_b90k & 45.34\% & 62.77\% & 15.77\% & 21.50\% & 79.55\% & 59.30\% \\
ViT-L-14 & commonpool\_xl\_s13b\_b90k & 43.50\% & 62.09\% & 14.49\% & 21.65\% & 76.41\% & 58.67\% \\
ViT-L-14-quickgelu & metaclip\_400m & 42.91\% & 61.60\% & 12.90\% & 21.43\% & 78.62\% & 59.11\% \\
ViT-L-14-quickgelu & metaclip\_fullcc & 43.02\% & 61.92\% & 13.70\% & 21.93\% & 81.63\% & 59.22\% \\
ViT-L-14-quickgelu & dfn2b & 46.37\% & 60.27\% & 16.88\% & 20.65\% & 80.08\% & 58.98\% \\
ViT-H-14 & laion2b\_s32b\_b79k & 45.32\% & 63.44\% & 15.51\% & 21.35\% & 79.06\% & 60.24\% \\
ViT-H-14-quickgelu & metaclip\_fullcc & 43.12\% & 62.44\% & 13.65\% & 21.72\% & 81.02\% & 59.99\% \\
ViT-H-14-quickgelu & dfn5b & 46.39\% & 60.99\% & 16.93\% & 21.21\% & 82.36\% & 60.19\% \\
ViT-g-14 & laion2b\_s12b\_b42k & 45.06\% & 61.44\% & 15.83\% & 21.22\% & 78.95\% & 58.44\% \\
ViT-g-14 & laion2b\_s34b\_b88k & 44.95\% & 63.72\% & 15.58\% & 22.15\% & 77.94\% & 60.40\% \\
ViT-bigG-14 & laion2b\_s39b\_b160k & 45.09\% & 62.14\% & 15.67\% & 20.48\% & 79.37\% & 60.44\% \\
roberta-ViT-B-32 & laion2b\_s12b\_b32k & 12.58\% & 47.85\% & 0.04\% & 14.40\% & 48.48\% & 47.37\% \\
xlm-roberta-base-ViT-B-32 & laion5b\_s13b\_b90k & 12.96\% & 50.62\% & 0.03\% & 16.57\% & 49.08\% & 50.49\% \\
xlm-roberta-large-ViT-H-14 & frozen\_laion5b\_s13b\_b90k & 11.99\% & 51.06\% & 0.01\% & 17.67\% & 50.67\% & 50.59\% \\
\end{tabular}
\caption{Summarized Dense Captions test results on OpenCLIP models. We compare various baseline models on our Subcrop-Caption Matching (SCM) and negatives tests.}
\label{tab:dci-full-test}
\end{table*}

\clearpage
\clearpage
%----------------------------------------
\section{DCI Dataset Datasheet}
\label{sec:datasheet}

The following are our answers to the Datasheets for Datasets~\citep{gebru2021datasheets} question list.

\subsection{Motivation}
\textbf{For what purpose was the dataset created?} To create an initial dataset of highly aligned text and image pairs that were not yet available, primarily for evaluating how well existing models can make use of all of the data. \\
\textbf{Who created the dataset and on behalf of which entity?} Researchers on Meta's FAIR research team created it on their own behalf.\\
%\textbf{Who created the dataset and on behalf of which entity?} Researchers on [Redacted for anonymity] created it on their own behalf.\\
\textbf{Who funded the creation of the dataset?} Meta
%\textbf{Who funded the creation of the dataset?} [Redacted for anonymity]

\subsection{Composition}
\textbf{What do the instances that comprise the dataset represent?} Images with text annotations \\
\textbf{How many instances are there in total?} 7805 images with complete mask-aligned annotations.\\
\textbf{Does the dataset contain all possible instances or is it a sample of instances from a larger set?} Images were sampled from a random subset of SA-1B's~\citep{kirillov2023segment} underlying image dataset.\\
\textbf{What data does each instance consist of?} One image, a top-level caption and description, and then a list of submask-subcaption pairings covering a significant portion of the image's content\\
\textbf{Is there a label or target associated with each instance?} Just the text descriptions, no categorization is done.\\
\textbf{Is any information missing from individual instances?} Not all of the image is covered in the submask-aligned captions, so the descriptions may still be considered incomplete. \\
\textbf{Are relationships between individual instances made explicit?} There is no clear relationships between instances in the dataset. \\
\textbf{Are there recommended data splits? } The dataset is intended \textbf{primarily} as a test set, however we also provide a finetuneing train/valid/test split for those wanting to use it for experiments.\\
\textbf{Are there any errors, sources of noise, or redundancies in the
dataset?} Though attempts were made to keep the dataset high-quality, annotator error can be present through the dataset, modeling errors may cause some masks to have been omitted, and the LLM-based augmentation for scaling captions to CLIP length may introduce noise as well.\\
\textbf{Is the dataset self-contained, or does it link to or otherwise rely on
external resources?} Self-contained\\
\textbf{Does the dataset contain data that might be considered confidential?} Not to the authors' knowledge\\
\textbf{Does the dataset contain data that, if viewed directly, might be offensive, insulting, threatening, or might otherwise cause anxiety?} Not to the author's knowledge

\subsection{Collection Process}
\textbf{How was the data associated with each instance acquired?} The images were selected from the SA-1B image dataset, and underwent a combination of automated and manual annotation\\
\textbf{What mechanisms or procedures were used to collect the data?} We use the Mephisto framework, as well as custom annotation interfaces, to collect the data. Complete details are available on the project's GitHub.\\
\textbf{If the dataset is a sample from a larger set, what was the sampling
strategy?} Random selection from a single subset\\
\textbf{Who was involved in the data collection process and how were they compensated?} Crowdworkers were paid well above minimum wage for their time spent.\\
\textbf{Over what timeframe was the data collected?} Spring through Fall of 2023. \\
\textbf{Were any ethical review processes conducted?} This collection process underwent internal review.

\subsection{Preprocessing/cleaning/labeling}
\textbf{Was any preprocessing/cleaning/labeling of the data done?} The dataset was preprocessed using the Segment Anything Model in order to identify the regions of the image to be annotated.\\
\textbf{Was the “raw” data saved in addition to the preprocessed/cleaned/labeled
data?} The raw data is unmodified by the extraction process.\\
\textbf{Is the software that was used to preprocess/clean/label the data
available?} All of the software used to construct the dataset will be made available alongside the dataset release on the project's GitHub.

\subsection{Uses}
\textbf{Has the dataset been used for any tasks already?} The dataset is used for the Densely Captioned Images test set. \\
\textbf{Are there tasks for which the dataset should not be used?} The dataset is intended as a test set. Any use outside of this is unplanned by the authors. \\

\subsection{Distribution}
\textbf{Will the dataset be distributed to third parties outside of the entity on behalf of which the dataset was created?} The dataset will be made broadly available \\
\textbf{How will the dataset will be distributed?} A download script will be made available on the project GitHub, alongside a copy of the code used to collect and prepare the original dataset.\\
\textbf{When will the dataset be distributed?} Upon release of the associated publication.\\
\textbf{Will the dataset be distributed under a copyright or other intellectual property (IP) license, and/or under applicable terms of use
(ToU)?} The dataset will be released under CC-By-NC.\\
\textbf{Have any third parties imposed IP-based or other restrictions on
the data associated with the instances?} Not to the authors' knowledge.\\
\textbf{Do any export controls or other regulatory restrictions apply to
the dataset or to individual instances?} Not to the authors' knowledge.\\

\subsection{Maintenance}
\textbf{Who will be supporting/hosting/maintaining the dataset?} Meta's FAIR team will host this dataset.\\
%\textbf{Who will be supporting/hosting/maintaining the dataset?} [Redacted for anonymity] will host this dataset.\\
\textbf{How can the owner/curator/manager of the dataset be contacted?} On the project's GitHub page. \\
\textbf{Is there an erratum?} Changes will be noted on the project's github.\\
\textbf{Will the dataset be updated?} The authors have no clear schedule to update or alter the dataset.\\
\textbf{If the dataset relates to people, are there applicable limits on the
retention of the data associated with the instances?} All known instances of people in the dataset have been face-blurred as per the SA-1B release, and no retention policy is known.\\
\textbf{If others want to extend/augment/build on/contribute to the
dataset, is there a mechanism for them to do so?} They may do so from the project GitHub, following the terms included therein.\\

\end{document}